# Distributed Stochastic Multi-Task Learning with Graph Regularization

Weiran Wang[*], Jialei Wang[†], Mladen Kolar[‡], and Nathan Srebro[*]

[*]Toyota Technological Institute at Chicago, IL, USA
[†]Department of Computer Science, University of Chicago, IL, USA
[‡]Booth School of Business, University of Chicago, IL, USA

## Abstract

We propose methods for distributed graph-based multi-task learning that are based on weighted averaging of messages from other machines. Uniform averaging or diminishing stepsize in these methods would yield consensus (single task) learning. We show how simply skewing the averaging weights or controlling the stepsize allows learning different, but related, tasks on the different machines.

## 1 Introduction

We consider a distributed learning problem in a multi-task setting: each machine $i$ has access to samples from a different data distribution $\mathcal{D}_i$, with potentially a different optimal predictor, and thus a different learning task, but where we still assume some similarity between different tasks. The goal of each machine is to find a good predictor for its own task, based on its own local data, as well as communicating with the other machines so as to leverage the similarity to other related tasks.

Distributed multi-task learning lies between a homogeneous distributed learning setting (e.g. Shamir and Srebro, 2014), where all machines have data from the same source distribution, and inhomogeneous consensus problems (e.g. Ram et al., 2010; Boyd et al., 2011; Balcan et al., 2012), where each machine sees data from a different source, but the goal is to reach a single consensus predictor. In many distributed learning problems, different machines do indeed see different distributions. For example, machines might serve different geographical regions. In a more extreme "federated learning" (Konecny et al., 2015) scenario, each machine is a single user device, and its data distribution might reflect e.g. the user's speech, language biases, usage patterns, etc. Such heterogeneity requires departing from a homogeneous model. But if the data distribution on each machine is different, we might as well learn a personalized predictor for each machine, while still leveraging commonalities as in multi-task learning, instead of insisting on consensus. Unlike when seeking consensus, we could learn a predictor entirely locally, ignoring data on other machines. But the premise of multi-task learning is that by communicating with other machines we can improve our predictions, reduce the sample complexity, and hopefully also reduce the computational cost on each machine by distributing the computation.

Central to multi-task learning is the notion of relatedness between tasks. In a high-dimensional setting, with large number of variables, we might expect a small common set of predictive variables, where the form of the dependence on variables in this common set varies between tasks (Turlach et al., 2005; Obozinski et al., 2011; Lounici et al., 2011; Wang et al., 2015). Another approach is to assume that the predictors lie in a shared lower dimensional subspace (Ando and Zhang, 2005; Yuan et al., 2007; Wang et al., 2016) or all have low-norm under some shared linear representation (Amit et al., 2007; Argyriou et al., 2008). Both the shared sparsity and shared subspaces models have recently been considered in a distributed learning setting (Wang et al., 2015, 2016), and nuclear-norm regularized multi-task learning has been studied from a distributed optimization perspective (Baytas et al., 2016).

In this paper, we consider graph-based multi-task learning, where relatedness between tasks is specified through a weighted graph over the tasks. Neighboring tasks in the graph are expected to be similar, with a penalty for dis-similarity specified by the weight between them (see precise formulation in Section 2) (Maurer, 2006; Evgeniou et al., 2005). This also generalized a simpler "fully connected" multi-task model where all predictors are close to each other (Evgeniou and Pontil, 2004). A predictor-homogeneous assumption can also be viewed as an extreme case where all weights go to infinity, forcing all predictors to be identical. In distributed multi-task learning, graph-based relatedness is especially appealing if the relatedness graph also matches the graph of network links between machines, as might be the case, e.g. in a geographical setting or with physical sensors. We therefor emphasize and prefer methods with communication only between neighboring tasks on the graph.

In designing methods for graph-based multi-task learning, we are interested in methods that (1) are natural and simple—all our algorithms have a similar and natural structure, involving weighted averaging of messages from neighboring machines and a local gradient or prox calculation; (2) have low communication costs, are sample efficient, and preferably also have low computational cost; and (3) are backed by rigorous guarantees on the amount of communication, samples and computation required.

Graph-based multi-task learning has been recently studied by Vanhaesebrouck et al. (2017) and Liu et al. (2017), both considering the problem as distributed optimization of the multi-task regularized *empirical* objective, similar to our approach in Section 3.2). Vanhaesebrouck et al. suggested an asynchronous gossip-type algorithms and an ADMM procedure, while Liu et al. proposed using SDCA, and also considered learning the relatedness graph itself. Neither provides any statistical analysis, nor analysis of the iteration complexity and communication cost based on the methods. We conduct detailed comparison of convergence properties with these methods in Appendix H, providing upper bounds of their iteration complexities when possible; our methods have faster convergence than the guarantees we could obtain for them. Also, neither directly considers the underlying learning problem (minimizing the actual expected errors), and so neither studies stochastic methods (in the flavor of our Section 4).

Here, we show how methods that arise naturally by skewing averaging weights or controlling stepsize of consensus learning methods do yield good guarantees. We also propose stochastic methods which allow reducing the computational cost, and we compare the empirical performance of both our batch and stochastic methods to those of Vanhaesebrouck et al. (2017) and Ma et al. (2015).

**Notations** In this paper, boldface lower-case letters denote column vectors, boldface capital letters denote matrices, $vec(\mathbf{U})$ is the vectorial form of a matrix $\mathbf{U}$ which concatenates columns of $\mathbf{U}$, and $\mathbf{U} \otimes \mathbf{V}$ is the Kronecker product between two matrices $\mathbf{U}$ and $\mathbf{V}$. Furthermore, $\langle \mathbf{u}, \mathbf{v} \rangle = \mathbf{u}^\top \mathbf{v}$ denotes the inner product of two vectors $\mathbf{u}$ and $\mathbf{v}$, while $\langle \mathbf{U}, \mathbf{V} \rangle = \operatorname{tr}\left(\mathbf{U}^\top \mathbf{V}\right)$ denotes inner product of two matrices $\mathbf{U}$ and $\mathbf{V}$ of the same dimensions. We use $\|\mathbf{u}\| = \sqrt{\langle \mathbf{u}, \mathbf{u} \rangle}$ to denote the length of a vector $\mathbf{u}$, $\|\mathbf{U}\|_F = \|vec(\mathbf{U})\|$ the Frobenius norm of a matrix $\mathbf{U}$, and $\|\mathbf{U}\|_\mathbf{M} = \sqrt{\operatorname{tr}\left(\mathbf{U}\mathbf{M}\mathbf{U}^\top\right)} = \sqrt{\langle \mathbf{U}\mathbf{M}, \mathbf{U} \rangle}$ the norm of $\mathbf{U}$ with respect to some positive definite matrix $\mathbf{M}$. A function $f(\mathbf{x})$ is Lipschitz if $|f(\mathbf{x}) - f(\mathbf{y})| \leq L \|\mathbf{x} - \mathbf{y}\|$, $\forall \mathbf{x}, \mathbf{y}$. A convex function $f(\mathbf{x})$ is $\beta$-smooth and $\mu$-strongly convex if $\frac{\mu}{2} \|\mathbf{x} - \mathbf{y}\|^2 \leq f(\mathbf{x}) - f(\mathbf{y}) - \langle \nabla f(\mathbf{y}), \mathbf{x} - \mathbf{y} \rangle \leq \frac{\beta}{2} \|\mathbf{x} - \mathbf{y}\|^2$, $\forall \mathbf{x}, \mathbf{y}$. This definition extends to functions of matrices, by replacing the vector norm with the Frobenius norm in the above inequality.

## 2 Graph-based multi-task learning

Consider a distributed setting with $m$ machines, where each machine $i$ has access to a data distribution $\mathcal{D}_i$ and would like to learn a predictor $\mathbf{w}_i \in \mathbb{R}^d$ for each machines with small expected loss $F_i(\mathbf{w}_i) = \mathbb{E}_{\mathbf{z}_i \sim \mathcal{D}_i} [\ell(\mathbf{w}_i, \mathbf{z}_i)]$. A known weighted graph, with known non-negative weights $\{a_{ik}\}$, specifies the relatedness between tasks. Specially, we would like to consider predictor matrices $\mathbf{W} = [\mathbf{w}_1, \mathbf{w}_2, \ldots, \mathbf{w}_m] \in \mathbb{R}^{d \times m}$ from the set

$$\Omega = \left\{ \mathbf{W} : \|\mathbf{w}_i\|^2 \leq B^2, \quad \forall i = 1, \ldots, m, \right.$$
$$\left. \sum_{i \neq k} \frac{a_{ik}}{2} \|\mathbf{w}_i - \mathbf{w}_k\|^2 \leq S^2 \right\},$$

i.e., we would like the norm of each individual predictor to be bounded (so that it has low complexity and generalizes well), and the weighted dis-similarities between related predictors to also be small.

Taking an agnostic PAC-learning approach, our goal is to minimize the overall *population* objective

$$F(\mathbf{W}) := \frac{1}{m} \sum_{i=1}^m \mathbb{E}_{\mathbf{z}_i \sim \mathcal{D}_i} [\ell(\mathbf{w}_i, \mathbf{z}_i)], \tag{1}$$

and be competitive with respect to predictors in the set $\Omega$. Denoting $\mathbf{W}^* = \arg\min_{\mathbf{W} \in \Omega} F(\mathbf{W})$ the optimal predictor from $\Omega$, and we would like to learn a predictor $\mathbf{W}$ with $F(\mathbf{W}) \leq F(\mathbf{W}^*) + \varepsilon$.

In our analysis, we take the instantaneous loss $\ell(\mathbf{w}, \mathbf{z})$ to be $L$-Lipschitz continuous, and sometimes also assume it is smooth. In the latter case, we assume machine $i$'s loss $\ell(\mathbf{w}_i, \mathbf{z}_i)$ is $\beta_i$-smooth in $\mathbf{w}_i$, and so the global loss $\widehat{F}(\mathbf{W})$ is $\left(\frac{\beta_F}{m}\right)$-smooth in $\mathbf{W}$ with $\beta_F = \max_{i=1,\ldots,m} \beta_i$. Even ignoring the constraint on the similarity between predictors, the sample complexity for each individual task (i.e. the number of samples from $\mathcal{D}_i$ required to ensure $F_i(\mathbf{w}_i) \leq F_i(\mathbf{w}_i^*) + \varepsilon$) is $n_L = \mathcal{O}\left(\frac{L^2 B^2}{\epsilon^2}\right)$. That is, with a total of $\mathcal{O}\left(\frac{m L^2 B^2}{\epsilon^2}\right)$ samples, we can learn $\mathbf{W}$ with the desired guarantee $F(\mathbf{W}) \leq F(\mathbf{W}^*) + \varepsilon$ without any communication between the machines, by, e.g., solving an independent $\ell_2$-regularized ERM problem on each machine. This *local* approach is the baseline on which any method involving communication between the machines should improve.

**Graph Laplacian** The term $\sum_{i \neq k} \frac{a_{ik}}{2} \|\mathbf{w}_i - \mathbf{w}_k\|^2$ can be written equivalently using the graph Laplacian. Let $\mathbf{A} = [a_{ik}] \in \mathbb{R}^{m \times m}$ be the adjacency matrix, and $\mathbf{L} = \operatorname{diag}(\mathbf{A}\mathbf{1}) - \mathbf{A}$ be the corresponding graph Laplacian ($\mathbf{L}_{ik} = \sum_{l \neq i} a_{il}$ if $i = k$, and $\mathbf{L}_{ik} = -a_{ik}$ otherwise), so that $\sum_{i \neq k} \frac{a_{ik}}{2} \|\mathbf{w}_i - \mathbf{w}_k\|^2 = \sum_{i,k} \mathbf{L}_{ik} \langle \mathbf{w}_i, \mathbf{w}_k \rangle = \text{tr}(\mathbf{W}\mathbf{L}\mathbf{W}^\top)$. The eigenvalues of $\mathbf{L}$ will play an important role and we denote them by $0 = \lambda_1 \leq \cdots \leq \lambda_m$.

**Regularized ERM** One way for learning the predictors is to solve the regularized empirical risk minimization (ERM) problem. Let $\widehat{F}_i(\mathbf{w}_i) = \frac{1}{n}\sum_{j=1}^n \ell(\mathbf{w}_i, \mathbf{z}_{ij})$ be the local empirical loss of machine $i$, and let $Z = \{\mathbf{z}_{ij} : i = 1, \ldots, m, j = 1, \ldots, n\}$ be the sample set. The regularized ERM objective is

$$\widehat{\mathbf{W}} = \arg\min_{\mathbf{W}} \underbrace{\frac{1}{m}\sum_{i=1}^m \widehat{F}_i(\mathbf{w}_i)}_{\widehat{F}(\mathbf{W})} + \underbrace{\frac{\eta}{2m}\sum_{i=1}^m \|\mathbf{w}_i\|^2 + \frac{\tau}{2m}\text{tr}(\mathbf{W}\mathbf{L}\mathbf{W}^\top)}_{R(\mathbf{W})}, \quad (2)$$

where $\eta, \tau \geq 0$ are regularization parameters. Let $\widehat{\mathbf{W}} = \arg\min_{\mathbf{W}} \widehat{F}(\mathbf{W}) + R(\mathbf{W})$ be the solution to (2).

To understand the statistical property of multi-task learning and facilitate further discussion, we first analyze the generalization error of $\widehat{\mathbf{W}}$. Inspired by Maurer (2006), who showed essentially the same learning guarantee for the solution of a *constrained* ERM problem (i.e., $\arg\min_{\mathbf{W} \in \Omega} \widehat{F}(\mathbf{W})$), we provide guarantee for the *regularized* ERM solution $\widehat{\mathbf{W}}$. Our motivation for studying regularized ERM rather than constrained ERM is that it is easier to solve unconstrained problem using (proximal) gradient methods, and we avoid computing projection onto the constraint set $\Omega$, which is difficult in a distributed setting.[1]

While the analysis of Maurer (2006) was based on the Rademacher complexity of $\Omega$ (and required the solution to lie in $\Omega$), our proof uses the stability based argument for generalization with strongly convex regularizers (Shalev-Shwartz et al., 2009). Our analysis also reveals a fundamental connection between single- and multi-task learning: to obtain generalization of a single task in the distributed setting, we only need concentration for the sampling process of that task. In our case, we consider strong convexity w.r.t. the $\|\mathbf{W}\|_\mathbf{M}$-norm where $\mathbf{M} = \mathbf{I} + \frac{\tau}{\eta}\mathbf{L}$.

**Lemma 1.** *Assume that the instantaneous loss $\ell(\mathbf{w}, \mathbf{z})$ is $L$-Lipschitz with respect to $\mathbf{w}$. Then for the ERM solution defined in (2), we have $\mathbb{E}_Z\left[F(\widehat{\mathbf{W}}) - \widehat{F}(\widehat{\mathbf{W}})\right] \leq \frac{4L^2}{mn}\sum_{i=1}^m \frac{1}{\eta + \tau\lambda_i}$.*

**Corollary 2.** *Set $\eta = \frac{2LB\sqrt{\frac{1+m\cdot\rho(B,S)}{mn}}}{B^2}$ and $\tau = \frac{2LB\sqrt{\frac{1+m\cdot\rho(B,S)}{mn}}}{S^2/m}$ in (2), where*

$$\rho(B,S) := \frac{1}{m}\sum_{i=2}^m \frac{1}{1 + \lambda_i m B^2/S^2}.$$

*Then $\mathbb{E}_Z\left[F(\widehat{\mathbf{W}}) - F(\mathbf{W}^*)\right] \leq 4LB\sqrt{\frac{1+m\cdot\rho(B,S)}{mn}}$.*

The quantity $\rho(B, S)$ measures task relatedness and thus the benefit of multi-task learning. It depends on the parameters $(B, S)$ and the graph, but not the data. The value of $\rho(B, S)$ ranges from 0 (when $\lambda_i m B^2 \gg S^2$) to $\frac{m-1}{m} \leq 1$ (when $\lambda_i m B^2 \ll S^2$), corresponding to two extreme cases.

---

[1] Although for convex optimization, the constrained form and the regularized form are equivalent due to the Lagrange duality, solving the constrained form may still require repeatedly solving the regularized form and searching for the Lagrange multiplier.

- When $S$ is small and the graph is connected with high weights, the predictors are encouraged to be similar to each other (we have a consensus problem if $S = 0$ and the graph is connected), and $\rho(B,S)$ is close to 0. The generalization error is then $\mathcal{O}\left(\frac{LB}{\sqrt{mn}}\right)$, corresponding to that of single task learning using $mn$ samples.

- When $S$ is large or the graph is disconnected, tasks are not very related and $\rho(B,S)$ is close to 1. In this case, the generalization error behaves like $\mathcal{O}\left(\frac{LB}{\sqrt{n}}\right)$, and we are essentially performing local learning with $n$ samples for each task.

For a fixed number of machines $m$ and graph Laplacian $\mathbf{L}$, to achieve $\varepsilon$ excess population error by the above approach, the number of samples used by each machines is $n_C = \mathcal{O}\left(\frac{L^2B^2(1/m+\rho(B,S))}{\varepsilon^2}\right) = \mathcal{O}\left((1/m + \rho(B,S)) \cdot n_L\right)$. Therefore, when the tasks are related and $\rho(B,S)$ is small, the sample complexity of multi-task learning is significantly smaller than $n_L$ needed by the local approach.

To implement the regularized ERM approach in the distributed setting, we could have each machines send $n_C$ samples to a central machine, and then minimize the regularized empirical loss on that machine. We refer to this baseline as the *centralized* approach—it is sample efficient, but expensive in terms of communication and computation. We are interested in distributed multi-task learning algorithms that are also sample efficient, i.e. use only $O(n_C)$ samples on each machine (or at least, not much more then this), but have low computation and communication costs. This can be done either by low-communication distributed optimization of the regularized empirical error (2).

## 3 Distributed algorithms for ERM

In this section, we propose efficient distributed algorithms for minimizing the regularized empirical objective (2). The simplest approach is perhaps to perform gradient descent on $\widehat{F}(W)$. Interestingly, such updates take the form:

$$\mathbf{w}_i^{t+1} = \sum_{k=1}^{m} \mu_{ki}^{t+1} \mathbf{w}_k^t - \alpha^{t+1} \nabla \widehat{F}_i(\mathbf{w}_i^t), \tag{3}$$

where $\alpha^{t+1} > 0$ is the stepsize at iteration $t+1$, and the weights for combining neighboring predictors are

$$\mu_{ki}^{t+1} = \begin{cases} 1 - \alpha^{t+1}(\eta + \tau \sum_{k'} a_{ik'}) & : \text{if } i = k, \\ \alpha^{t+1} \tau a_{ik} & : \text{otherwise}. \end{cases} \tag{4}$$

With an appropriate step-size schedule (or even a fixed stepsize if the loss is smooth), this method converges to $\widehat{\mathbf{W}}$. Furthermore, the updates require only communication along the relatedness graph, since the update for each machines involves only predictors from neighboring machines (with nonzero affinities). This is already a very natural and intuitive method for distributed multi-task learning, and we will return to it later. When the loss is smooth, the method can be accelerated using Nesterov's techniques (Nesterov, 2004, as detailed in Appendix C) without any increase in communication costs nor substantial increase in computation. But first, we suggest two more powerful alternatives.

Taking steps based on the gradients amounts to considering, in each iteration, a linearization of the objective, that is of both the empirical loss $\widehat{F}(\mathbf{W})$ and the regularizer $R(\mathbf{W})$. However, in order to obtain a distributable update, it is sufficient to linearize only one of these components while treating the other more explicitly, since each one of them separately can be efficiently optimized in a distributed way: the empirical loss $\widehat{F}(\mathbf{W})$ decomposes over machines, and so can be directly optimized in a distributed way, while $R(\mathbf{W})$ is data independent and could be optimized implicitly based on the common knowledge of the relatedness graph. In the following, we consider two distributed schemes, each based on directly handling one of the components, and each preferable in a different regime depending on the relatedness graph and the structure and cost of communication.

## 3.1 Directly solving the regularizer

We first consider methods which directly handle the regularization term $R(\mathbf{W})$. To do so, we consider the change of variable $\mathbf{U}^t = \mathbf{W}^t \mathbf{M}^{\frac{1}{2}}$ where $\mathbf{M} = \mathbf{I} + \frac{\tau}{\eta}\mathbf{L}$, we can rewrite the ERM objective as

$$\min_{\mathbf{U}} \ \widehat{F}(\mathbf{U}\mathbf{M}^{-\frac{1}{2}}) + \frac{\eta}{2m}\|\mathbf{U}\|_F^2. \tag{5}$$

We propose to optimize this objective using gradient descent with respect to $\mathbf{U}$, which reduces to the updates in the $\mathbf{W}$-space: for $t = 0, \ldots,$

$$\mathbf{W}^{t+1} = \left(1 - \alpha^{t+1}\eta\right)\mathbf{W}^t - \alpha^{t+1}\nabla\widehat{F}(\mathbf{W}^t)\cdot \mathbf{M}^{-1} \tag{6}$$

where $\alpha^{t+1} > 0$ is the stepsize at iteration $t+1$. In each iteration, machine $i$ performs the following update with $\mu_{ki}^{t+1} = \alpha^{t+1}(\mathbf{M}^{-1})_{ki}$:

$$\mathbf{w}_i^{t+1} = \left(1 - \alpha^{t+1}\eta\right)\mathbf{w}_i^t - \sum_{k=1}^{m} \mu_{ki}^{t+1}\nabla\widehat{F}_k(\mathbf{w}_k^t). \tag{7}$$

This update can be implemented in the distributed setting with a broadcast channel: it requires that each machine has access to gradients of all machines, which can be achieved using one round of global, all-to-all communication (not respecting the graph). We could compute $\mathbf{M}^{-1}$ offline ahead of time, and need not re-calculated at each iteration.

When the loss is smooth, we can accelerate (7) using Nesterov's techniques without additional communication costs. Setting a constant stepsize $\frac{1}{\alpha^{t+1}} = \beta_F + \eta$, which is the smoothness parameter of the objective (5) in $\mathbf{U}^2$, to achieve $\epsilon$-suboptimality in (2), the iteration complexity of the accelerated algorithm is $\mathcal{O}\left(\sqrt{\frac{\beta_F + \eta}{\eta}}\log\frac{1}{\epsilon}\right)$. To achieve $\varepsilon$ excess error in the population loss, we set the optimization error $\epsilon = \mathcal{O}(\varepsilon)$ and plug in the choice of $\eta$ from Corollary 2, yielding the iteration complexity $\widetilde{\mathcal{O}}\left(\sqrt{\beta_F B^2/\varepsilon}\right)$.

## 3.2 Directly optimizing the loss

The above algorithm requires dense, broadcast communication for solving the proximal step defined by the graph. In a decentralized setting, it is desired to develop algorithms which use only local, peer-to-peer communication. This can be achieved by the updates below, where we linearize the graph regularizer but fully optimize over the loss:

$$\mathbf{W}^{t+1} = \arg\min_{\mathbf{W}} \langle \nabla R(\mathbf{W}^t), \mathbf{W} - \mathbf{W}^t \rangle$$
$$+ \frac{1}{2m\alpha^{t+1}} \|\mathbf{W} - \mathbf{W}^t\|_F^2 + \widehat{F}(\mathbf{W}), \qquad (8)$$

where $\alpha^{t+1}$ is the stepsize at iteration $t+1$. As (8) decouples over machines, machine $i$ independently computes a proximal operation using local data:

$$\mathbf{w}_i^{t+1} = \arg\min_{\mathbf{u}}$$
$$\frac{1}{2\alpha^{t+1}} \|\mathbf{u} - (\mathbf{w}_i^t - m\alpha^{t+1} \nabla_{\mathbf{w}_i} R(\mathbf{W}^t))\|^2 + \widehat{F}_i(\mathbf{u}).$$

---

[2]This is because $\nabla^2_{vec(\mathbf{U})}\widehat{F}(\mathbf{U}\mathbf{M}^{-\frac{1}{2}}) = (\mathbf{M}^{-\frac{1}{2}} \otimes \mathbf{I}) \cdot \nabla^2_{vec(\mathbf{W})}\widehat{F}(\mathbf{W}) \cdot (\mathbf{M}^{-\frac{1}{2}} \otimes \mathbf{I})$, and $\|\nabla^2_{vec(\mathbf{U})}\widehat{F}(\mathbf{U}\mathbf{M}^{-\frac{1}{2}})\| \leq \|\mathbf{M}^{-\frac{1}{2}}\| \cdot \|\nabla^2_{vec(\mathbf{W})}\widehat{F}(\mathbf{W})\| \cdot \|\mathbf{M}^{-\frac{1}{2}}\| \leq \frac{\beta_F}{m}$.

By the optimality condition of this update, we have

$$\mathbf{w}_i^{t+1} = \sum_{k=1}^{m} \mu_{ki}^{t+1} \mathbf{w}_k^t - \alpha^{t+1} \nabla \widehat{F}_i(\mathbf{w}_i^{t+1}), \qquad (9)$$

where the weights for combining neighboring predictors are the same as those in (4). Comparing (9) with the similar update (3) where we linearized both the regularizer and the loss, we observe that (9) is also a form of gradient method, with the gradient of loss evaluated at the "future" point.

The advantage of (9) is that the gradient $\nabla R(\mathbf{W})$ is data-independent and is obtained using only one round of local communication from each machine to its neighbors. Furthermore, the computation decouples over machines, and each machine optimizes the nonlinearized loss without communication. In fact, we need not solve the proximal steps exactly since the (accelerated) proximal gradient method is tolerant to errors in the steps (Schmidt et al., 2011), and sufficiently accurate solutions can often be obtained in time nearly linear in the number of examples processed using variance-reduced finite-sum methods such as SVRG (Johnson and Zhang, 2013). Overall, this is a communication-efficient approach in which each machine tries to spend significant amount of time performing local computations on its own data, and to communicate only infrequently. Note that similar proximal type operations also appear in the ADMM algorithm of Vanhaesebrouck et al. (2017), but the decoupling of tasks is different, because in the local problems of ADMM, each machine optimizes over also a copy of neighboring predictors.

We can again accelerate (9) using Nesterov's techniques, and set $\frac{1}{m\alpha^{t+1}} = \beta_R = \frac{\eta + \tau \lambda_m}{m}$, which is the smoothness parameter of $R(\mathbf{W})$ in $\mathbf{W}$. Then, to achieve $\varepsilon$ excess error in the population objective, the number of iterations needed by the accelerated algorithm is $\widetilde{\mathcal{O}}\left(\sqrt{\frac{\beta_R}{\eta/m}}\right) = \widetilde{\mathcal{O}}\left(\sqrt{\frac{\lambda_m m B^2}{S^2}}\right)$, using the choice of $\eta$ and $\tau$ from Corollary 2. We also show that this algorithm is tolerant to delay and analyze its convergence under bounded delay in Appendix G.

## 4  Stochastic algorithms

In ERM, we collect training samples on each machine ahead of time, and solve a fixed optimization problem defined by them. But in real-world scenarios, we might have access to virtually unlimited data, or a constantly available stream of examples. In this case, it might be statistically wasteful to reuse examples over iterations. Or, even if we do have a finite amount of data, as we shall see, we can get the same communication and statistical guarantee while processing only a minibatch at a time, thus significantly reducing computational cost. We consider stochastic variants of the approaches in Section 3 to directly optimize the population loss $F(\mathbf{W})$, using fresh samples in each update.

Table 1: Algorithms for distributed stochastic multi-task learning with graph regularization. Here $\varepsilon$ is the excess error in the population objective; $n_C = \mathcal{O}\left(\frac{L^2 B^2 \cdot (1/m + \rho(B,S))}{\varepsilon^2}\right)$ and $n_L = \mathcal{O}\left(\frac{L^2 B^2}{\epsilon^2}\right)$; $|\mathbf{E}|$ denotes the number of edges in the graph. For simplicity, schematic updates ignores acceleration, but the rates are given for the accelerated algorithms. Each cell shall be interpreted as $\widetilde{\mathcal{O}}(\cdot)$ which hides poly-logarithmic dependencies.

| Algorithms | Communication rounds | Vectors ($\in \mathbb{R}^d$) communicated per machine | Sample complexity per machine | Total Samples processed per machine |
|---|---|---|---|---|
| local | 0 | 0 | $n_L = \frac{n_C}{\frac{1}{m} + \rho(B,S)}$ | $n_L$ |
| centralized | | $n_C$ | $n_C = n_L \cdot (\frac{1}{m} + \rho(B,S))$ | $m \cdot n_C$ |
| ERM: directly solving regularizer<br>1. $\mathbf{g}_i^{t+1} = \sum_k \mu_{ki}^{t+1} \nabla \widehat{F}_k(\mathbf{w}_k^t)$<br>where $\mu_{ki}^{t+1} = \alpha^{t+1}(\mathbf{M}^{-1})_{ki}$<br>2. $\mathbf{w}_i^{t+1} = \mathbf{w}_i^t - \mathbf{g}_i^{t+1}$ | $\sqrt{\frac{B^2}{\varepsilon}}$ | $m \cdot \sqrt{\frac{B^2}{\varepsilon}}$ | $n_C$ | $n_C \cdot \sqrt{\frac{B^2}{\varepsilon}}$<br>$= n_C \cdot \sqrt[4]{n_L}$ |
| ERM: directly optimizing loss<br>1. $\widetilde{\mathbf{w}}_i^t = \sum_k \mu_{ki}^{t+1} \mathbf{w}_k^t$<br>where $\mu_{ki}^{t+1} = (\mathbf{I} - \alpha^{t+1}\eta\mathbf{M})_{ki}$<br>2. $\mathbf{w}_i^{t+1} = \widetilde{\mathbf{w}}_i^t - \alpha^{t+1} \nabla \widehat{F}_i^{t+1}(\mathbf{w}_i^{t+1})$ | $\sqrt{\frac{\lambda_m m B^2}{S^2}}$ | $\frac{|E|}{m} \cdot \sqrt{\frac{\lambda_m m B^2}{S^2}}$ | $n_C$ | $n_C \cdot \sqrt{\frac{\lambda_m m B^2}{S^2}}$ |
| Stochastic: directly solving regularizer<br>Algorithm 2, $b = \mathcal{O}\left(n_C \sqrt{\frac{\varepsilon}{B^2}}\right)$<br>1. $\mathbf{g}_i^{t+1} = \sum_k \mu_{ki}^{t+1} \nabla \widehat{F}_k^{t+1}(\mathbf{w}_k^t)$<br>where $\mu_{ki}^{t+1} = \alpha^{t+1}(\mathbf{M}^{-1})_{ki}$<br>2. $\mathbf{w}_i^{t+1} = \mathbf{w}_i^t - \mathbf{g}_i^{t+1}$ | $\sqrt{\frac{B^2}{\varepsilon}}$ | $m \cdot \sqrt{\frac{B^2}{\varepsilon}}$ | $n_C$ | $n_C$ |
| Stochastic: directly optimizing loss<br>1. $\widetilde{\mathbf{w}}_i^t = \sum_k \mu_{ki}^{t+1} \mathbf{w}_k^t$<br>where $\mu_{ki}^{t+1} = (\mathbf{I} - \alpha^{t+1}\eta\mathbf{M})_{ki}$<br>2. $\mathbf{w}_i^{t+1} = \widetilde{\mathbf{w}}_i^t - \alpha^{t+1} \nabla \widehat{F}_i^{t+1}(\mathbf{w}_i^{t+1})$ | | $\frac{|E|}{m}$ per iteration | $n_S$, probably $\in (n_C, n_L)$ | $n_S$ |

## 4.1 Directly solving the regularizer

Analogous to (7), we could perform minibatch SGD with $b$ samples per machine to approximate the gradient of the population loss: for $t = 0, \ldots,$

$$\mathbf{w}_i^{t+1} = \mathbf{w}_i^t - \sum_{k=1}^{m} \mu_{ki}^{t+1} \nabla \widehat{F}_k^{t+1}(\mathbf{w}_k^t). \tag{10}$$

where $\widehat{F}_k^{t+1}(\mathbf{w}_k^t) = \frac{1}{b} \sum_{j=1}^{b} \ell(\mathbf{w}_k^t, \mathbf{z}_{kj}^{t+1})$, and $\left\{\mathbf{z}_{kj}^{t+1} : j = 1, \ldots, b\right\}$ are $b$ samples drawn by machine $k$ at iteration $t+1$.

We can accelerate (10) using the accelerated stochastic approximation (AC-SA) algorithm of Lan (2012). We provide the detailed accelerated algorithm in both the **U**-space and **W**-space in Algorithm 2 (Appendix D). We have the following guarantee after running it for $T$ iterations.

**Theorem 3.** *Set the initialization $\mathbf{W}_0 = \mathbf{0}$ and stepsizes $\theta^{t+1} = \frac{t+1}{2}$, $\alpha^{t+1} = \frac{t+1}{2} \min\left\{\frac{m}{2\beta_F}, \frac{\sqrt{12mB^2}}{(T+2)^{\frac{3}{2}}\sigma}\right\}$ in Algorithm 2. Then $\mathbb{E}\left[F(\mathbf{W}_{ag}^T) - F(\mathbf{W}^*)\right] \leq \mathcal{O}\left(\frac{\sigma\sqrt{mB^2}}{\sqrt{bT}} + \frac{\beta_F B^2}{T^2}\right)$.*

**Sample complexity** Let $n = bT$ be the number of samples used in Algorithm 2. According to Theorem 3, as long as the minibatch size $b \leq b^* = \mathcal{O}\left(n\sqrt{\frac{\varepsilon(m,n)}{\beta_F B^2}}\right)$, the first term in the error bound is dominant and we achieve the generalization error $\mathcal{O}\left(\frac{\sigma\sqrt{mB^2}}{\sqrt{n}}\right) = \mathcal{O}\left(LB\sqrt{\frac{1+m\cdot\rho(B,S)}{mn}}\right)$ as in ERM, so we are still sample efficient in the stochastic setting.

**Time complexity** Algorithm 2 processes the drawn samples only once. While maintaining the sample efficiency, we can set the minibatch size to the largest value $b = b^*$, and this leads to the total number of iterations (and local communication rounds) $T^* = \frac{n}{b^*} = \mathcal{O}\left(\sqrt{\frac{\beta_F B^2}{\varepsilon(m,n)}}\right)$, also matching that of ERM. However, since each stochastic gradient uses only $b = o(n)$ samples, the local computation $\nabla \widehat{F}^{t+1}(\mathbf{W}^t)$ is significantly reduced.

## 4.2 Directly optimizing the loss

Analogous to (8), we can use the stochastic algorithm where at iteration $t+1$, machine $i$ computes

$$\mathbf{w}_i^{t+1} = \arg\min_{\mathbf{u}} \frac{1}{2\alpha^{t+1}} \left\|\mathbf{u} - \left(\mathbf{w}_i^t - m\alpha^{t+1}\nabla_{\mathbf{w}_i} R(\mathbf{W}^t)\right)\right\|^2 \\ + \frac{1}{b}\sum_{j=1}^{b} \ell(\mathbf{u}, \mathbf{z}_{ij}^{t+1}). \tag{11}$$

For $b = n$, it has the same per iteration computation cost as the ERM counterpart (both process $n$ samples in each iteration). But, intuitively, it would outperform the ERM algorithm for the same number of iterations/communications because it uses more fresh samples. We can prove the convergence of this algorithm, but do not have a satisfactory analysis showing it is sample efficient. We conjecture that its sample complexity per machine, denoted by $n_S$, is in the range $(n_C, n_L)$. We implemented the accelerated version of this simple algorithm and this conjecture seems to be supported by our experiments. In Appendix E, we provide a more complicated algorithm based on the minibatch-prox algorithm of Wang et al. (2017), that is sample efficient and trade off communication and memory costs.

**Comparison of the different approaches** Table 1 summarizes the communication and computation complexities of the proposed algorithms. Some of our methods require solving local regularized-ERM type problems on each machine. We do not analyze the precise complexity and required accuracy of such local computation, but keep track of the number of samples processed on each machine, i.e. sum of the sizes of the subproblems over the iterations, as the proxy for computational complexity. We emphasize that, despite the simplicity of our ERM methods, their have faster convergence than what we could obtain for previous methods; see detailed discussions in Appendix H. Our stochastic algorithms mirror the ERM algorithms in terms of updates, but can be computationally much more efficient.

## 5  Connection to consensus learning

The iterations we consider all involve taking a weighted average of messages (iterates or gradients) from other machines and a local gradient or prox computation. These same type of iterates have also been suggested and studied as methods for solving the consensus problem—that is, finding a single consensus predictor $\mathbf{w}$ that is good for all machines and minimizes $F(\mathbf{W}) = \frac{1}{m}\sum_{i=1}^{m} F_i(\mathbf{w}_i)$. But the consensus problem is fundamentally different from our "pluralistic" multi-task problem, with a different optimum. In this section we will understand what makes the same form of updates, namely updates of the form (3), (7), (9) or their stochastic variants, converge to either the consensus solution or to the pluralistic multi-task solution. In particular, we show how consensus methods are obtained as special cases of these updates, or as limits of the multi-task approach.

**Averaging gradients**  Let us begin with the update of the form (7) or its stochastic variant (10), where we take a weighted average of gradients from other machines. When the averaging weights are uniform, i.e. $\mu_{ki}^t = \alpha^t/m$ for all $i, k$, and as long as all machines start from the same initialization (e.g. $\mathbf{w}_i^t = 0$), the iterates will continue to be identical across machines throughout optimization (i.e. we will have $\mathbf{w}_i^t = \mathbf{w}_j^t$ for all $i, j, t$), thus maintaining consensus. Furthermore, the update (7) then boils down to precisely gradient descent on the empirical consensus objective $\widehat{F}(\mathbf{W}) + \frac{\eta}{2m}\|\mathbf{W}\|_F^2$, while the stochastic variant (7) is precisely a mini-batch stochastic gradient descent update on the consensus objective, with a mini-batch consisting of the union of the samples used across machines. Indeed, mini-batch SGD is a common approach for solving the distributed consensus problem, or for distributed learning in a homogeneous setting (where we assume the same distribution across machines, or at least the same good predictor). What we saw in Section 3, is that by changing to non-uniform weights, given by $\mu \propto \mathbf{M}^{-1}$, we can allow pluralism and converge to the multi-task solution.

We can furthermore observe how uniform weights (and therefor gradient descent/mini-batch SGD on the consensus problem) are obtained as a limit of the multi-task weights $\mu \propto \mathbf{M}^{-1}$. If the graph is connected, $\lambda_1 = 0$ is the only zero eigenvalue of the Laplacian $L$ with an associated eigenvector of $\mathbf{u} = [1, \ldots, 1]$ (if the graph is not connected, we cannot expect consensus, as each connected component will behave independently). Therefor $\mathbf{M}^{-1} = (\mathbf{I} + \frac{\tau}{\eta}\mathbf{L})^{-1}$ has a leading eigenvalue of 1 of multiplicity one, associated with the eigenvector $\mathbf{u}$. As $S \to 0$ and so $\tau \to \infty$, that is we are demanding increasing similarity between machines, the leading eigenvalue of $\mathbf{M}^{-1}$ remains 1 while all other eigenvalues go to zero, implying that $\mathbf{M}^{-1} \to \frac{1}{m}\mathbf{u}\mathbf{u}^\top$ and so $\mu_{ki}^t = \alpha^t \mathbf{M}_{ki}^{-1} \to \alpha^t/m$. That is, as we demand increasing similarity between machines, and thus converge to a consensus situation, the updates converge to standard consensus gradient descent or mini-batch SGD updates.

**Averaging iterates**  Let us now turn to updates of the form (3), the related prox updates (9), and their stochastic variants. Nedić and Ozdaglar (2009) proposed updates precisely of the form (3) as a decentralized procedure for the consensus problem. They showed that when the averaging weights $\mu_{ki}^t$ are doubly stochastic and do not vary between iterations (i.e. $\mu_{ki}^t = \mu_{ki}, \forall_k \sum_i \mu_{ki} = 1$ and $\forall_j \sum_k \mu_{ki} = 1$), and the stepsize on the gradient goes to zero, i.e. $\alpha^t \xrightarrow{t \to \infty} 0$, the updates (3) converge to the consensus solution. In our case, the averaging weights, as defined in (4), deviate from double-stochasticity, since $\sum_k \mu_{ki}^t = 1 - \alpha^t \eta$. Furthermore, and possibly more significantly, to obtain our convergence guarantees for smooth loss, we do not take $\alpha^t$ to zero. Even if we were to use diminishing stepsizes in our derivations, we would have $\alpha^t \to 0$, but in that case the averaging weights would not be fixed over iterations (as is the case in consensus optimization) and we would have $\mu^t \to \mathbf{I}$.

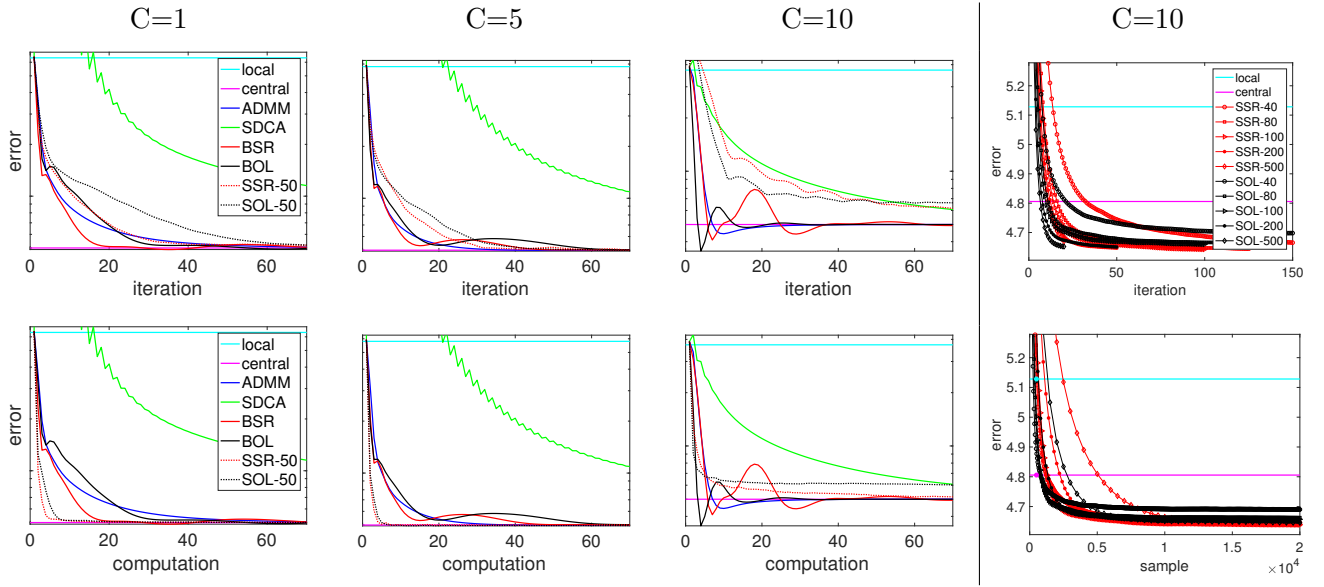

Figure 1: Results for regularized ERM (left panel) and our stochastic methods with different $b$ (right panel).

To see how consensus updates are obtained as a limiting case of our multi-task setting, we again consider a connected graph and study what happens as $S \to 0$ and so $\tau \to \infty$, while $B$ and therefor $\eta$ remain fixed. This corresponds to a fixed amount of local regularization, and increasing expectation that neighboring nodes are similar. Under this scaling, we would indeed have $\alpha = 1/(\eta + \tau \lambda_m) \to 0$, where $\lambda_m > 0$ since the graph is connected. Furthermore, we have that $\alpha \eta \to 0$ while $\alpha \tau \to 1/\lambda_m > 0$. Plugging this scaling into the multi-task averaging weights (4), we obtain the doubly stochastic weights:

$$\mu_{ki}^t \to \begin{cases} 1 - \frac{1}{\lambda_m} \sum_{k'} a_{ik'} & : \text{if } i = k, \\ \frac{1}{\lambda_m} a_{ik} & : \text{otherwise.} \end{cases} \qquad (12)$$

To summarize, a significant differentiation between consensus and multi-task learning is therefor in whether $\alpha^t$ diminishes *relative to* $(\mu^t - \mathbf{I})$. When our relatedness constraints approach consensus, $\alpha^t$ can diminish while $\mu^t$ is non-trivial and doubly stochastic. In fact, in studying consensus optimization, Yuan et al. (2016) recently noted that when $\alpha^t$ does not diminish, the methods does not converge to the consensus solution but only to a neighborhood of it. In light of our analysis, we now understand that this "neighborhood" corresponds to the multi-task learning solution, which indeed becomes increasingly similar to the consensus solution as $S \to 0$.

**Connection to the decentralized algorithm of Scaman et al. (2017)** When the graph is connected, the consensus constraint $\mathbf{w}_1 = \cdots = \mathbf{w}_m$ can be equivalently written as $\mathbf{W}\sqrt{\mathbf{L}} = \mathbf{0}$, since the null space of $\mathbf{L}$ contains only vectors of constants. Then the multi-task formulation (2) is a relaxation of

$$\min_{\mathbf{W}\sqrt{\mathbf{L}}=\mathbf{0}} \frac{1}{m}\sum_{i=1}^m \widehat{F}_i(\mathbf{w}_i) + \frac{\eta}{2m}\sum_{i=1}^m \|\mathbf{w}_i\|^2 \tag{13}$$

with the quadratic term $\frac{\tau}{2}\operatorname{tr}(\mathbf{WLW}^\top)$ penalizing the constraint violation. The quadratic penalty $\frac{\tau}{2}\operatorname{tr}(\mathbf{WLW}^\top)$ may lead to a large condition number for our algorithm (8) as $\tau \to \infty$.

Recently, Scaman et al. (2017) proposed an algorithm with optimal iteration/communication complexities for decentralized consensus learning, which performs accelerated gradient descent on the dual problem of (13), with updates (before acceleration):

$$\begin{aligned}
\mathbf{W}^{t+1} &= \arg\max_{\mathbf{W}} \langle \mathbf{V}^t, \mathbf{W}\rangle - \widehat{F}(\mathbf{W}), \\
\mathbf{V}^{t+1} &= \mathbf{V}^t - \alpha \mathbf{W}^{t+1}\mathbf{L},
\end{aligned} \tag{14}$$

where $\mathbf{V}^0 = \mathbf{0}$ and $\alpha > 0$ is the stepsize. It can be seen that their algorithm consists of the same type of basic operations (weighted local average of predictors, and solutions of local subproblems involving non-linearized loss) as ours. As noted by the authors, this is a form of distributed augmented Lagrangian method without the quadratic penalty.

## 6 Experiments

We examine different graph-based multi-task learning methods on the task of least squares regression using synthetic data. More details of the experiments (including data generation and more results) are given in Appendix I. The tasks are grouped into $C$ clusters and the true predictors within the same cluster are generated from the same Gaussian distribution, thus smaller $C$ implies higher task relatedness. We have input dimension $d = 100$, number of tasks $m = 100$, training set size $n = 500$, and vary number of task clusters $C$ over $\{1, 5, 10, 50\}$. We also generate a dev set of 10000 samples per task for tuning hyper-parameters, and test set of 10000 samples per task for approximately evaluating the population loss. The affinity graph $\mathbf{A} \in \mathbb{R}^{100 \times 100}$ is a (connected) 10-nearest neighbor graph with binary weights built on the true predictors.

The methods compared here are: `Local`, which solves a local ERM problem (with $\ell_2$-regularization) with $n$ samples for each task; `Centralized`, which solves the regularized ERM problem (2) with $n$ samples for each task; `ADMM`, which is the synchronized version of the algorithm of Vanhaesebrouck et al. (2017); `SDCA`, which is the algorithm used by Liu et al. (2017) for fixed graph; our algorithms are denoted as B/S (batch/stochastic) + SR/OL (solve regularizer/optimize loss).

**Empirical risk minimization** We fist compare the iterative methods on the regularized ERM problem (2), to which the analysis for `ADMM` and `SDCA` applies. We tune the $\ell_2$ regularization parameter for `Local` and $(\eta, \tau)$ for `Centralized`, and then fix the optimal $(\eta, \tau)$ for other methods. We also tune the quadratic penalty parameter for `ADMM`, the task separability and stepsize parameters for `SDCA`, and stepsize parameter for BSR/BOL (although the default value based on the smoothness parameter already works well for them). For SSR/SOL, we draw random samples from the fixed training set (with size $n$), and simply fix the minibatch size to be $n/10$.

Figure 1 (left panel) shows for each method the estimated $F(\mathbf{W})$ over iterations (or rounds of communication) in the top row, and over the amount of computation (measured by the number of passes over the training set) in the bottom row. Observe that all iterative algorithms converge to the same ERM solution, our algorithms tend to consistently outperform `ADMM` and `SDCA`.

**Stochastic optimization** We next demonstrate the efficiency of true stochastic algorithms (using fresh samples for each update) at $C = 10$. We allow the algorithms to process a total of 10000 fresh samples on each machine, and vary the minibatch size $b$ over $\{40, 80, 100, 200, 500\}$. The parameters $(\eta, \tau)$ are fixed to those used in the ERM experiments.

Figure 1 (right panel) shows for each method the estimated $F(\mathbf{W})$ over iterations (or rounds of communication) in the left plot, and over the amount of fresh samples processed (or total computation cost) in the right plot. As a reference, the error of `Local` and `Centralized` (using $n = 500$ samples per machine) are also given in the plots. We observe that with fresh samples, stochastic algorithms are competitive to ERM algorithms in terms of sample complexity, while being computationally more efficient.

## A  Proof of Lemma 1

Recall that the ERM problem is defined as

$$\widehat{\mathbf{W}} = \arg\min_{\mathbf{W}} \ \widehat{F}(\mathbf{W}) + R(\mathbf{W}) := \frac{1}{m} \sum_{i=1}^m \widehat{F}_i(\mathbf{w}_i) + \frac{\eta}{2m} \sum_{i=1}^m \|\mathbf{w}_i\|^2 + \frac{\tau}{2m} \operatorname{tr}\left(\mathbf{W}\mathbf{L}\mathbf{W}^\top\right),$$

where $\eta, \tau \geq 0$ are regularization parameters, $Z = \{\mathbf{z}_{ij} : i = 1, \ldots, m, j = 1, \ldots, n\}$ is the sample set. And recall that $\lambda_i$, $i = 1, \ldots, m$ are the eigenvalues of $\mathbf{L}$.

Assume that the instantaneous loss $\ell(\mathbf{w}, \mathbf{z})$ is $L$-Lipschitz in $\mathbf{w}$. We would like to show that

$$\mathbb{E}_Z \left[ F(\widehat{\mathbf{W}}) - \widehat{F}(\widehat{\mathbf{W}}) \right] \leq \frac{4L^2}{mn} \sum_{i=1}^m \frac{1}{\eta + \tau \lambda_i}.$$

*Proof.* In the following, we define $\mathbf{M} = \mathbf{I} + \frac{\tau}{\eta}\mathbf{L}$ which is positive definite. Furthermore, perform the following change of variables

$$\mathbf{U} = \mathbf{W}\mathbf{M}^{\frac{1}{2}}, \quad \Leftrightarrow \quad \mathbf{w}_i = (\mathbf{U}\mathbf{M}^{-\frac{1}{2}}) \cdot \mathbf{e}_i$$

where $\mathbf{e}_i$ is the $i$-th standard basis in $\mathbb{R}^m$.

We can then rewrite the losses using the new variables:

$$\frac{1}{m}\ell(\mathbf{w}_i, \mathbf{z}_i) = \frac{1}{m}\ell(\mathbf{U}\mathbf{M}^{-\frac{1}{2}}\mathbf{e}_i, \mathbf{z}_i) =: h_i(\mathbf{U}, \mathbf{z}_i), \quad \text{for} \quad i = 1, \ldots, m,$$

and the empirical objective as

$$\min_{\mathbf{U} \in \mathbb{R}^{d \times m}} \ \widehat{H}(\mathbf{U}) := \frac{1}{n} \sum_{j=1}^n h_1(\mathbf{U}, \mathbf{z}_{1j}) + \left( \sum_{i=2}^m \frac{1}{n} \sum_{j=1}^n h_i(\mathbf{U}, \mathbf{z}_{ij}) + \frac{\eta}{2m} \|\mathbf{U}\|_F^2 \right). \tag{15}$$

We can view (15) as performing ERM in the space of $\mathbf{U}$, using the instantaneous loss $h_1(\mathbf{U}, \mathbf{z}_1)$ with $n$ independent samples $\{\mathbf{z}_{1j}\}_{j=1,\ldots,n}$, and using the term in bracket as the $\mathbf{z}_1$-independent regularizer.

Recall that the ERM solution to an objective with Lipschitz loss and strongly convex regularizer is stable. Obviously, the regularization term in (15) is $\left(\frac{\eta}{m}\right)$-strongly convex in $\mathbf{U}$. We now bound the Lipschitz constant of $h_1(\mathbf{U}, \mathbf{z}_1)$ in $\mathbf{U}$. Observe that

$$\nabla_{\mathbf{U}} h_1(\mathbf{U}, \mathbf{z}_1) = \frac{1}{m} \nabla_{\mathbf{w}_i} \ell(\mathbf{w}_1, \mathbf{z}_1) \cdot \mathbf{e}_1^\top \mathbf{M}^{-\frac{1}{2}},$$

and as a result the Lipschitz constant is bounded by

$$\|\nabla_{\mathbf{U}} h_1(\mathbf{U}, \mathbf{z}_1)\|_F = \frac{1}{m} \sqrt{\operatorname{tr}\left(\nabla_{\mathbf{w}_1} \ell(\mathbf{w}_1, \mathbf{z}_1) \cdot \mathbf{e}_1^\top \mathbf{M}^{-1} \mathbf{e}_1 \cdot \nabla_{\mathbf{w}_1} \ell(\mathbf{w}_1, \mathbf{z}_1)^\top\right)} \leq \frac{L \sqrt{(\mathbf{M}^{-1})_{11}}}{m}$$

where we have used the $L$-Lipschitz continuity of $\ell(\mathbf{w}_1, \mathbf{z}_1)$ which implies $\|\nabla_{\mathbf{w}_1} \ell(\mathbf{w}_1, \mathbf{z}_1)\| \leq L$.

According to Shalev-Shwartz et al. (2009)[Theorem 6], for any fixed $\{\mathbf{z}_{ij}\}_{\substack{i=2,\ldots,m \\ j=1,\ldots,n}}$, it holds for the ERM solution $\widehat{\mathbf{U}} = \arg\min_{\mathbf{U}} \widehat{H}(\mathbf{U}) = \widehat{\mathbf{W}} \mathbf{M}^{\frac{1}{2}}$ that

$$\left| \mathbb{E}_{\{\mathbf{z}_{1j}\}} \left[ \mathbb{E}_{\mathbf{z}_1}[h_1(\widehat{\mathbf{U}}, \mathbf{z}_1)] - \frac{1}{n} \sum_{j=1}^n h_1(\widehat{\mathbf{U}}, \mathbf{z}_{1j}) \right] \right| \leq 4 \left( \frac{L \sqrt{(\mathbf{M}^{-1})_{11}}}{m} \right)^2 \bigg/ (\eta n/m) = \frac{4 L^2 (\mathbf{M}^{-1})_{11}}{\eta m n}.$$

Translating this in terms of the original variables, we have

$$\forall \{\mathbf{z}_{ij}\}_{\substack{i=2,\ldots,m \\ j=1,\ldots,n}}, \qquad \left| \mathbb{E}_{\{\mathbf{z}_{1j}\}} \left[ F_1(\widehat{\mathbf{W}}) - \widehat{F}_1(\widehat{\mathbf{W}}) \right] \right| \leq \frac{4 L^2 (\mathbf{M}^{-1})_{11}}{\eta n}$$

where $F_1(\mathbf{W}) = \mathbb{E}_{\mathbf{z}_1}[\ell(\mathbf{w}_1, \mathbf{z}_1)]$ and $\widehat{F}_1(\mathbf{W}) = \frac{1}{n} \sum_{j=1}^n \ell(\mathbf{w}_1, \mathbf{z}_{1j})$.

By the convexity of $|\cdot|$ and the Jensen's inequality, this implies

$$\left| \mathbb{E}_Z \left[ F_1(\widehat{\mathbf{W}}) - \widehat{F}_1(\widehat{\mathbf{W}}) \right] \right| = \left| \mathbb{E}_{\{\mathbf{z}_{ij}\}_{\substack{i=2,\ldots,m \\ j=1,\ldots,n}}} \left[ \mathbb{E}_{\{\mathbf{z}_{1j}\}} \left[ F_1(\widehat{\mathbf{W}}) - \widehat{F}_1(\widehat{\mathbf{W}}) \right] \right] \right|$$

$$\leq \mathbb{E}_{\{\mathbf{z}_{ij}\}_{\substack{i=2,\ldots,m \\ j=1,\ldots,n}}} \left| \mathbb{E}_{\{\mathbf{z}_{1j}\}} \left[ F_1(\widehat{\mathbf{W}}) - \widehat{F}_1(\widehat{\mathbf{W}}) \right] \right| \leq \frac{4 L^2 (\mathbf{M}^{-1})_{11}}{\eta n}.$$

This result shows that, to obtain generalization for a single task, we only need concentration for the sampling process of that task. By the same argument, we obtain similar inequalities regarding stability for losses on each machine.

Finally, we have by the triangle inequality that

$$\left| \mathbb{E}_Z \left[ F(\widehat{\mathbf{W}}) - \widehat{F}(\widehat{\mathbf{W}}) \right] \right| \leq \frac{1}{m} \sum_{i=1}^m \left| \mathbb{E}_Z \left[ F_i(\widehat{\mathbf{W}}) - \widehat{F}_i(\widehat{\mathbf{W}}) \right] \right|$$

$$\leq \frac{1}{m} \sum_{i=1}^m \frac{4 L^2 (\mathbf{M}^{-1})_{ii}}{\eta n}$$

$$= \frac{4 L^2 \operatorname{tr}\left(\mathbf{M}^{-1}\right)}{\eta m n}$$

$$= \frac{4 L^2 \sum_{i=1}^m \frac{1}{1 + \tau \lambda_i / \eta}}{\eta m n}$$

which is what we set out to prove. □

## B   Proof of Lemma 2

Based on Lemma 1, we now show that by properly setting the regularization parameters in the regularized ERM problem (2), i.e., $\eta = \frac{2LB\sqrt{\frac{1+m\cdot\rho(B,S)}{mn}}}{B^2}$ and $\tau = \frac{2LB\sqrt{\frac{1+m\cdot\rho(B,S)}{mn}}}{S^2/m}$, we have that

$$\mathbb{E}_Z\left[F(\widehat{\mathbf{W}}) - F(\mathbf{W}^*)\right] \leq 4LB\sqrt{\frac{1 + m\cdot\rho(B,S)}{mn}}.$$

where $\rho(B,S) := \frac{1}{m}\sum_{i=2}^{m}\frac{1}{1+\lambda_i mB^2/S^2}$.

*Proof.* Observe that

$$\mathbb{E}_Z\left[F(\widehat{\mathbf{W}})\right] \leq \mathbb{E}_Z\left[\widehat{F}(\widehat{\mathbf{W}})\right] + \frac{4L^2}{mn}\sum_{i=1}^{m}\frac{1}{\eta + \tau\lambda_i}$$

$$\leq \mathbb{E}_Z\left[\widehat{F}(\widehat{\mathbf{W}}) + R(\widehat{\mathbf{W}})\right] + \frac{4L^2}{mn}\sum_{i=1}^{m}\frac{1}{\eta + \tau\lambda_i}$$

$$\leq \mathbb{E}_Z\left[\widehat{F}(\mathbf{W}^*) + R(\mathbf{W}^*)\right] + \frac{4L^2}{mn}\sum_{i=1}^{m}\frac{1}{\eta + \tau\lambda_i}$$

$$= F(\mathbf{W}^*) + R(\mathbf{W}^*) + \frac{4L^2}{mn}\sum_{i=1}^{m}\frac{1}{\eta + \tau\lambda_i}$$

where we have used Lemma 1 in the first inequality, and that $\widehat{\mathbf{W}}$ is the empiric risk minimizer in the third inequality.

Since $\mathbf{W}^* \in \Omega$, we can bound the excess error as

$$\varepsilon(m,n) = \mathbb{E}_Z\left[F(\widehat{\mathbf{W}}) - F(\mathbf{W}^*)\right] \leq R(\mathbf{W}^*) + \frac{4L^2}{mn}\sum_{i=1}^{m}\frac{1}{\eta + \tau\lambda_i}$$

$$\leq \frac{1}{2}\eta B^2 + \frac{1}{2m}\tau S^2 + \frac{4L^2}{mn}\sum_{i=1}^{m}\frac{1}{\eta + \tau\lambda_i}. \tag{16}$$

Now, set $\eta = \frac{\epsilon}{B^2}$ and $\tau = \frac{m\epsilon}{S^2}$ for some $\epsilon$ that will be specified later. Continuing from (16) yields

$$\varepsilon(m,n) \leq \epsilon + \frac{4L^2}{mn}\sum_{i=1}^{m}\frac{1}{\frac{\epsilon}{B^2} + \frac{m\epsilon}{S^2}\lambda_i}$$

$$= \epsilon + \frac{1}{\epsilon}\cdot\frac{4L^2B^2}{n}\cdot\frac{1}{m}\sum_{i=1}^{m}\frac{1}{1 + \lambda_i mB^2/S^2}$$

$$\leq \epsilon + \frac{1}{\epsilon}\cdot\left(\frac{4L^2B^2}{mn} + \frac{4L^2B^2}{n}\cdot\frac{1}{m}\sum_{i=2}^{m}\frac{1}{1 + \lambda_i mB^2/S^2}\right)$$

$$\leq \epsilon + \frac{1}{\epsilon}\cdot\left(\frac{4L^2B^2}{mn} + \frac{4L^2B^2}{n}\cdot\rho(B,S)\right).$$

Minimizing the RHS over $\epsilon$ gives $\epsilon = 2LB\sqrt{\frac{1}{mn} + \frac{\rho(B,S)}{n}}$, and

$$\varepsilon(m,n) \leq 4LB\sqrt{\frac{1}{mn} + \frac{\rho(B,S)}{n}}.$$

□

## C  The accelerated proximal gradient algorithm

We provide the accelerated proximal gradient algorithms in Algorithm 1, which are used to accelerate our ERM algorithms in the main text. The proximal operator is defined as $\text{prox}_h^\beta(\mathbf{x}) = \arg\min_{\mathbf{y}} \frac{\beta}{2} \|\mathbf{y} - \mathbf{x}\|^2 + h(\mathbf{y})$ where $\beta > 0$ and $h(\mathbf{x})$ is convex and possibly non-smooth.

---

**Algorithm 1** ProxGrad($g$,$h$,$\beta$,$\mu$): Accelerated proximal gradient descent.

---

**Input:** Objective has the form $f(\mathbf{w}) = g(\mathbf{w}) + h(\mathbf{w})$, where $g(\mathbf{w})$ is $\beta$-smooth and $\mu$-strongly convex, and $h(\mathbf{w})$ is convex.
  Initialize $\mathbf{w}^0$, $\mathbf{y}^1 \leftarrow \mathbf{w}^0$
  **for** $t = 1, \ldots, T$ **do**
    $\mathbf{w}^t \leftarrow \text{prox}_h^\beta\left(\mathbf{y}^t - \frac{1}{\beta}\nabla g(\mathbf{y}^t)\right)$, $\qquad \mathbf{y}^{t+1} \leftarrow \mathbf{w}^t + \frac{\sqrt{\beta}-\sqrt{\mu}}{\sqrt{\beta}+\sqrt{\mu}}\left(\mathbf{w}^t - \mathbf{w}^{t-1}\right)$
  **end for**
**Output:** $\mathbf{w}^T$ is the approximate solution.

---

## D  Analysis of stochastic optimization by directly solving the regularizer

In each iteration of this algorithm, we draw $b$ samples per machine to approximate the gradient of the population loss and perform minibatch SGD, which amounts to linearizing the loss on a minibatch. The key to being sample efficient is to respect the geometry imposed by the graph Laplacian.

As in Section 3.1, define the change of variable $\mathbf{U}^t = \mathbf{W}^t \mathbf{M}^{\frac{1}{2}}$ where $\mathbf{M} = \mathbf{I} + \frac{mB^2}{S^2}\mathbf{L}$. Our population objective is $F(\mathbf{W}) = F(\mathbf{U}\mathbf{M}^{-\frac{1}{2}})$, and the predictor $\mathbf{U}^* = \mathbf{W}^*\mathbf{M}^{\frac{1}{2}}$ satisfies the constraint that $\|\mathbf{U}^*\|_F^2 = \text{tr}\left(\mathbf{W}^*\left(\mathbf{I} + \frac{mB^2}{S^2}\mathbf{L}\right)(\mathbf{W}^*)^\top\right) \leq 2mB^2$. We can perform minibatch SGD in the $\mathbf{U}$-space:

$$\mathbf{U}^{t+1} = \arg\min_{\mathbf{U}} \quad \alpha^{t+1}\langle \nabla \widehat{F}^{t+1}(\mathbf{U}^t \mathbf{M}^{-\frac{1}{2}}) \cdot \mathbf{M}^{-\frac{1}{2}}, \mathbf{U} - \mathbf{U}^t\rangle + \frac{1}{2}\left\|\mathbf{U} - \mathbf{U}^t\right\|_F^2, \qquad \text{for} \quad t = 0, \ldots,$$

where $\widehat{F}^{t+1}(\mathbf{W}^t) = \frac{1}{mb}\sum_{i=1}^m \sum_{j=1}^b \ell(\mathbf{w}_i^t, \mathbf{z}_{ij}^{t+1})$ and $\left\{\mathbf{z}_{ij}^{t+1}\right\}_{j=1,\ldots,b}$ are $b$ samples drawn by machine $i$ at iteration $t+1$, and $\alpha^{t+1} > 0$ is a stepsize parameter. In the $\mathbf{W}$-space, the above update reduces to

$$\mathbf{W}^{t+1} = \mathbf{W}^t - \alpha^{t+1}\nabla \widehat{F}^{t+1}(\mathbf{W}^t) \cdot \mathbf{M}^{-1},$$

Clearly, this update requires inverting the graph Laplacian.

We can further accelerate this method using the accelerated stochastic approximation (AC-SA) algorithm of Lan (2012). We give the detailed stochastic algorithm by directly solving the regularizer (with linearized loss) in Algorithm 2.

---

**Algorithm 2** Accelerated minibatch SGD. This algorithm maintains three iterate sequences: $\{\mathbf{U}^t\}$ is the sequence of prox centers, $\{\mathbf{U}^t_{md}\}$ is the "middle" sequence with which we evaluate the stochastic gradient and build models (approximations) of the objective, and $\{\mathbf{U}^t_{ag}\}$ is the "aggregated" sequence with which we evaluate the objective values.

---

**Input:** The stepsize sequences $\{\theta^{t+1}\}$ and $\{\alpha^{t+1}\}$ for $t = 0, \ldots$.
   Initialize $\mathbf{W}^0 \leftarrow \mathbf{0}, \quad \mathbf{W}^0_{ag} \leftarrow \mathbf{W}^0$                                                $\{\mathbf{U}^0 \leftarrow \mathbf{0}, \quad \mathbf{U}^0_{ag} \leftarrow \mathbf{U}^0\}$
   **for** $t = 0, \ldots, T-1$ **do**
      $\mathbf{W}^t_{md} \leftarrow (\theta^{t+1})^{-1} \mathbf{W}^t + (1 - (\theta^{t+1})^{-1}) \mathbf{W}^t_{ag}$    $\{\mathbf{U}^t_{md} \leftarrow (\theta^{t+1})^{-1} \mathbf{U}^t + (1 - (\theta^{t+1})^{-1}) \mathbf{U}^t_{ag}\}$
      $\mathbf{W}^{t+1} \leftarrow \mathbf{W}^t - \alpha^{t+1} \nabla \widehat{F}^{t+1}(\mathbf{W}^t_{md}) \cdot \mathbf{M}^{-1}$
                                                          $\{\mathbf{U}^{t+1} \leftarrow \mathbf{U}^t - \alpha^{t+1} \nabla \widehat{F}^{t+1}(\mathbf{U}^t_{md} \mathbf{M}^{-\frac{1}{2}}) \cdot \mathbf{M}^{-\frac{1}{2}}\}$
      $\mathbf{W}^{t+1}_{ag} \leftarrow (\theta^{t+1})^{-1} \mathbf{W}^{t+1} + (1 - (\theta^{t+1})^{-1}) \mathbf{W}^t_{ag}$  $\{\mathbf{U}^{t+1}_{ag} \leftarrow (\theta^{t+1})^{-1} \mathbf{U}^{t+1} + (1 - (\theta^{t+1})^{-1}) \mathbf{U}^t_{ag}\}$
   **end for**
**Output:** $\mathbf{W}^T_{ag}$ (or equivalently $\mathbf{U}^T_{ag}$) is the approximate solution.

---

The key quantity for analyzing the convergence property of minibatch SGD is the variance of stochastic gradients in the $\mathbf{U}$-space, which we now derive. We can view $\xi = (\mathbf{z}_1, \ldots, \mathbf{z}_m)$ as the combined sample, $\ell_{multi}(\mathbf{W}, \xi) = \frac{1}{m} \sum_{i=1}^m \ell(\mathbf{w}_i, \mathbf{z}_i)$ as the averaged instantaneous loss, so that $\widehat{F}^{t+1}(\mathbf{W}) = \frac{1}{b} \sum_{j=1}^b \ell_{multi}(\mathbf{W}, \xi_j^{t+1})$ approximates $\mathbb{E}_\xi[\ell_{multi}(\mathbf{W}, \xi)]$ with $b$ combined samples. The lemma below bounds the variance of stochastic gradient estimated with one combined sample.

**Lemma 4.** *The variance of stochastic gradient in the $\mathbf{U}$-space is bounded:*

$$\mathbb{E}_\xi \left\| \nabla \ell_{multi}\left(\mathbf{U}\mathbf{M}^{-\frac{1}{2}}, \xi\right) \cdot \mathbf{M}^{-\frac{1}{2}} - \mathbb{E}_\xi \left[ \nabla \ell_{multi}\left(\mathbf{U}\mathbf{M}^{-\frac{1}{2}}, \xi\right) \cdot \mathbf{M}^{-\frac{1}{2}} \right] \right\|_F^2 \leq \sigma^2$$

*where* $\sigma^2 := \frac{4L^2}{m^2}(1 + m \cdot \rho(B, S))$.

*Proof.* By direct calculation, we have

$$\mathbb{E}_\xi \left\| \nabla \ell_{multi}\left(\mathbf{U}\mathbf{M}^{-\frac{1}{2}}, \xi\right) \cdot \mathbf{M}^{-\frac{1}{2}} - \mathbb{E}_\xi \left[ \nabla \ell_{multi}\left(\mathbf{U}\mathbf{M}^{-\frac{1}{2}}, \xi\right) \cdot \mathbf{M}^{-\frac{1}{2}} \right] \right\|_F^2$$

$$= \frac{1}{m^2} \mathbb{E}_\xi \| [\nabla_{\mathbf{w}_1} \ell(\mathbf{w}_1, \mathbf{z}_1) - \mathbb{E}_{\mathbf{z}_1}[\nabla_{\mathbf{w}_1} \ell(\mathbf{w}_1, \mathbf{z}_1)], \ldots, \nabla_{\mathbf{w}_m} \ell(\mathbf{w}_m, \mathbf{z}_m) - \mathbb{E}_{\mathbf{z}_m}[\nabla_{\mathbf{w}_m} \ell(\mathbf{w}_m, \mathbf{z}_m)]] \|^2_{\mathbf{M}^{-1}}$$

$$= \frac{1}{m^2} \sum_{i,k} \mathbb{E}_{\mathbf{z}_i, \mathbf{z}_k} \langle \nabla_{\mathbf{w}_i} \ell(\mathbf{w}_i, \mathbf{z}_i) - \mathbb{E}_{\mathbf{z}_i}[\nabla_{\mathbf{w}_i} \ell(\mathbf{w}_i, \mathbf{z}_i)], \nabla_{\mathbf{w}_k} \ell(\mathbf{w}_k, \mathbf{z}_k) - \mathbb{E}_{\mathbf{z}_k}[\nabla_{\mathbf{w}_k} \ell(\mathbf{w}_k, \mathbf{z}_k)] \rangle \cdot (\mathbf{M}^{-1})_{ik}$$

$$= \frac{1}{m^2} \sum_{i=1}^m \| \nabla_{\mathbf{w}_i} \ell(\mathbf{w}_i, \mathbf{z}_i) - \mathbb{E}_{\mathbf{z}_i}[\nabla_{\mathbf{w}_i} \ell(\mathbf{w}_i, \mathbf{z}_i)] \|^2 \cdot (\mathbf{M}^{-1})_{ii} \tag{17}$$

$$\leq \frac{4L^2}{m^2} \operatorname{tr}(\mathbf{M}^{-1}) = \frac{4L^2}{m^2} \sum_{i=1}^m \frac{1}{1 + \lambda_i m B^2 / S^2} = \frac{4L^2}{m^2}(1 + m \cdot \rho(B, S)) = \sigma^2$$

where we have used the independence between $\mathbf{z}_i$ and $\mathbf{z}_k$ for $i \neq k$ so that the cross terms vanishes in (17), and the triangle inequality and that $\|\nabla_{\mathbf{w}_i} \ell(\mathbf{w}_i, \mathbf{z}_i)\| \leq L$ in the inequality. □

Averaging the $b$ independent stochastic gradients on a minibatch reduces the gradient variance to $\sigma^2/b$ (see, e.g., Dekel et al., 2012, eqn 7). Note that $\frac{\beta_F}{m}$ is the smoothness parameter of $F(\mathbf{UM}^{-\frac{1}{2}})$ w.r.t. $\mathbf{U}$, and the distance generating function $\frac{1}{2}\|\mathbf{U}\|_F^2$ is 1-strongly convex w.r.t. the $\|\mathbf{U}\|_F$-norm. Plugging these problem parameters into (Lan, 2012)(Corollary 1) yields Theorem 3.

# E  A sample-efficient stochastic algorithm by directly optimizing the loss

The key to sample efficiency in the stochastic setting is to couple the individual learning tasks with the graph, and respect the geometry of the $\mathbf{U}$-space (e.g., in deriving the generalization performance in Lemma 1, we rely on strong convexity in the norm $\|\mathbf{U}\|_F$). This motivates us to derive a sample-efficient stochastic algorithm based on the minibatch-prox method (Wang et al., 2017). The minibatch-prox method solves a subproblem involving nonlinearized loss on a minibatch in each iteration, and was shown to have the optimal sample complexity for stochastic convex optimization regardless of the minibatch size (recall from Section 4.1 that mnibatch SGD achieves the optimal sample complexity only for small enough minibatch size), and it was the basis for developing communication- and memory-efficient algorithm for distributed stochastic consensus learning in Wang et al. (2017).

We detail the minibatch-prox based algorithm in Algorithm 3, which consists of two nested loops. In the outer loop, we perform minibatch-prox in the space of $\mathbf{U}$; in each iteration of the outer loop we use $b$ samples per machines to approximate the nonlinearized loss, and approximately solves a subproblem involving the full Laplacian in the $\mathbf{W}$-space. The solutions to the subproblems (which is then a small ERM problem with fixed samples) are computed approximately by the inner loops, where we perform acclerated gradient descent in the space of $\mathbf{W}$.

---
**Algorithm 3** Distributed minibatch prox.
---
Initialize $\mathbf{W}^0 \leftarrow \mathbf{0}$
for $t = 0, \ldots, T-1$ do
  Approximately solve
  $\quad \mathbf{W}^{t+1} \approx \widehat{\mathbf{W}}^{t+1} = \arg\min_{\mathbf{W}} \; \frac{\gamma}{2} \operatorname{tr}\left((\mathbf{W} - \mathbf{W}^t)\mathbf{M}(\mathbf{W} - \mathbf{W}^t)^\top\right) + \widehat{F}^{t+1}(\mathbf{W})$
  to $\zeta_{t+1}$-suboptimality using the accelerated proximal gradient algorithm
  $\quad \texttt{ProxGrad}(\gamma \operatorname{tr}\left((\mathbf{W} - \mathbf{W}^t)\mathbf{M}(\mathbf{W} - \mathbf{W}^t)^\top\right), \widehat{F}^{t+1}(\mathbf{W}), \gamma(1 + \frac{mB^2}{S^2}\lambda_m), \gamma)$
end for
**Output:** $\overline{\mathbf{W}} = \frac{1}{T}\sum_{t=1}^T \mathbf{W}^t$ is the approximate solution.

---

The minibatch-prox algorithm for minimizing $F(\mathbf{UM}^{-\frac{1}{2}})$ works as follows:

$$\mathbf{U}^{t+1} \approx \widehat{\mathbf{U}} = \arg\min_{\mathbf{U}} \; \frac{\gamma}{2}\left\|\mathbf{U} - \mathbf{U}^t\right\|_F^2 + \widehat{F}^{t+1}(\mathbf{UM}^{-\frac{1}{2}}), \qquad \text{for } t = 0, \ldots, \tag{18}$$

where in each iteration we draw $b$ fresh samples per machine to approximate $F(\mathbf{W})$ by $\widehat{F}^{t+1}(\mathbf{W}) = \frac{1}{mb}\sum_{i=1}^m \sum_{j=1}^b \ell(\mathbf{w}_i, \mathbf{z}_{ib}^{t+1})$. Note that we allow inexact solutions to the objective in (18). The corresponding update of (18) in the $\mathbf{W}$-space is $\mathbf{W}^{t+1} \approx \widehat{\mathbf{W}}^{t+1} = \arg\min_{\mathbf{W}} \; \widehat{f}^{t+1}(\mathbf{W})$ where

$$\widehat{f}^{t+1}(\mathbf{W}) = \frac{\gamma}{2}\operatorname{tr}\left((\mathbf{W} - \mathbf{W}^t)\mathbf{M}(\mathbf{W} - \mathbf{W}^t)^\top\right) + \widehat{F}^{t+1}(\mathbf{W}). \tag{19}$$

We provide the learning guarantee of the minibatch-prox algorithm in the following theorem.

**Theorem 5.** *Suppose that we initialize Algorithm 3 with $\mathbf{W} = \mathbf{0}$ and set $\gamma = 2\sqrt{\frac{T}{b}} \cdot \frac{L\sqrt{1+m\cdot\rho(B,S)}}{m^{\frac{3}{2}}B}$. Assume that for all $t \geq 0$, the error in minimizing (19) satisfies*

$$\widehat{f}^{t+1}(\mathbf{W}^{t+1}) - \min_{\mathbf{W}} \widehat{f}^{t+1}(\mathbf{W}) \leq \zeta_{t+1} = \min\left(\left(\frac{T}{b}\right)^{\frac{1}{2}}, \left(\frac{T}{b}\right)^{\frac{3}{2}}\right) \cdot \frac{LB(1 + m \cdot \rho(B,S))^{\frac{3}{2}}}{m^{\frac{5}{2}}t^3}.$$

*Then for $\overline{\mathbf{W}}^T = \frac{1}{T}\sum_{t=1}^{T} \mathbf{W}^t$, we have $\mathbb{E}\left[F(\overline{\mathbf{W}}^T) - F(\mathbf{W}^*)\right] = \mathcal{O}\left(\frac{LB\sqrt{1+m\cdot\rho(B,S)}}{\sqrt{mbT}}\right).$*

*Proof.* Let $L_{\mathbf{U}} = \frac{L\sqrt{\operatorname{tr}(M^{-1})}}{m}$ where $\operatorname{tr}(\mathbf{M}^{-1}) = 1 + m \cdot \rho(B, S)$. By an analysis similar to that of Lemma 1 (and essentially due to $\widehat{f}^{t+1}(\mathbf{W})$'s $\gamma$-strong convexity w.r.t. the norm $\|\cdot\|_{\mathbf{M}}$), we obtain the "stability" of the exact minimizer to (19), i.e., $\left|\mathbb{E}[\widehat{F}^{t+1}(\widehat{\mathbf{W}}^{t+1}) - F(\widehat{\mathbf{W}}^{t+1})]\right| \leq \frac{4L^2\operatorname{tr}(\mathbf{M}^{-1})}{\gamma m^2 b} = \frac{4L_{\mathbf{U}}^2}{\gamma b}$. Furthermore, if the suboptimality of $\mathbf{W}^{t+1}$ satisfies $\widehat{f}^{t+1}(\mathbf{W}^{t+1}) - \widehat{f}^{t+1}(\widehat{\mathbf{W}}^{t+1}) \leq \zeta_{t+1}$, by the $\gamma$-strong convexity of $\widehat{f}^{t+1}(\mathbf{W})$ w.r.t. the Euclidean norm, we have

$$\|\mathbf{w}_i^{t+1} - \widehat{\mathbf{w}}_i^{t+1}\| \leq \sqrt{\frac{2\zeta_{t+1}}{\gamma}}, \qquad \text{for } i = 1, \ldots, m,$$

and consequently by the Lipschitz continuity of the loss, we have

$$\widehat{F}^{t+1}(\mathbf{W}^{t+1}) - \widehat{F}(\widehat{\mathbf{W}}^{t+1}) \leq \sqrt{\frac{2L^2\zeta_{t+1}}{\gamma}} = \sqrt{\frac{2L_{\mathbf{U}}^2}{\gamma} \cdot \frac{m^2\zeta_{t+1}}{\operatorname{tr}(M^{-1})}}.$$

This reconstructs the essential lemma required by the minibatch-prox analysis (Wang et al., 2017, Lemma 2). We can then invoke the learning guarantee of minibatch-prox (Wang et al., 2017, Theorem 7), by using our $L_{\mathbf{U}}$ in place of their $L$, and our $\frac{m^2\zeta_{t+1}}{\operatorname{tr}(M^{-1})}$ in place of their $\eta_t$.

In the end, we have

$$\mathbb{E}\left[F(\overline{\mathbf{W}}^T) - F(\mathbf{W}^*)\right] \leq \mathcal{O}\left(\frac{LB\sqrt{\operatorname{tr}(\mathbf{M}^{-1})}}{\sqrt{mbT}}\right) = \mathcal{O}\left(\frac{LB\sqrt{1+m\cdot\rho(B,S)}}{\sqrt{mbT}}\right).$$

$\square$

For fixed $n = bT$, minibatch-prox attains the generalization error $\mathcal{O}\left(LB\sqrt{\frac{1+m\cdot\rho(B,S)}{mn}}\right)$ for any minibatch size $b$. Though the error in solving each subproblem (19) seems stringent as it decreases over iterations, we can apply the linearly convergent accelerated proximal gradient method in the inner loops to the subproblems. For any minibatch size $b$, the number of outer iterations is $T = \frac{n}{b}$, and the number of inner iterations for each outer iteration (the initial error for the subproblems are bounded with a warm-start, see Appendix F) is $\widetilde{\mathcal{O}}\left(\sqrt{\frac{\lambda_m m B^2}{S^2}}\right)$, so the total number of communication rounds is the multiplication $\widetilde{\mathcal{O}}\left(\frac{n}{b} \cdot \sqrt{\frac{\lambda_m m B^2}{S^2}}\right)$.

This algorithm allows us to trade off communication and memory: We could use small number of samples $b$ in each outer iteration (limited by the local memory), but the total number communication rounds increase with $\frac{1}{b}$. The most communication-efficient setting is $b = n$, in which case we are essentially solving one ERM problem with $mn$ samples (by linearzing the regularizer). Finally, we note that each update of the simple algorithm (11) (without the outer+inner loop structure) and a single inner iteration of the minibatch-prox subproblem (19) have the same communication/computation costs.

## F  Warm start when directly optimizing the loss

**Lemma 6.** *Consider the objective of the proximal operator*

$$\min_{\mathbf{y}} \ f(\mathbf{y}) = \frac{\beta}{2} \|\mathbf{y} - \mathbf{x}\|^2 + h(\mathbf{y}).$$

*where $h(\mathbf{y})$ is $L$-Lipschitz, and let $\mathbf{x}^* = \arg\min_{\mathbf{y}} \ f(\mathbf{y})$. Then we have*

$$\|\mathbf{x}^* - \mathbf{x}\| \leq L/\beta,$$

*and the suboptimality of $\mathbf{x}$ is bounded*

$$f(\mathbf{x}) - f(\mathbf{x}^*) \leq L^2/\beta.$$

*Proof.* By the first-order optimality of $\mathbf{x}^*$, we have

$$\mathbf{0} = \beta(\mathbf{x}^* - \mathbf{x}) + \nabla h(\mathbf{x}^*)$$

where $\nabla h(\mathbf{x}^*)$ is a subgradient of $h$ at $\mathbf{x}^*$. By the assumption that $h(\mathbf{y})$ is $L$-Lipschitz, we have $\|\nabla h(\mathbf{x}^*)\| \leq L$ and consequently $\|\mathbf{x}^* - \mathbf{x}\| = \|\nabla h(\mathbf{x}^*)\|/\beta \leq L/\beta$.

For the suboptimality of $\mathbf{x}$, it follows again from the Lipschitz continuity of $h$ that

$$\begin{aligned} f(\mathbf{x}) - f(\mathbf{x}^*) &= 0 + h(\mathbf{x}) - \frac{\beta}{2}\|\mathbf{x}^* - \mathbf{x}\|^2 - h(\mathbf{x}^*) \\ &\leq h(\mathbf{x}) - h(\mathbf{x}^*) \\ &\leq L \|\mathbf{x} - \mathbf{x}^*\| \\ &\leq L^2/\beta. \end{aligned}$$

$\square$

This lemma indicates that for solving the local objectives when directly optimizing the loss, e.g., (8), we can initialize from $\mathbf{W}^t - \frac{1}{\beta}\nabla R(\mathbf{W}^t)$ which mixes the local predictor with those of the neighbors, and the initial suboptimality of this warm start is bounded by $\mathcal{O}\left(\frac{L^2}{\beta_F}\right)$.

A similar result holds when the distance term is defined by other non-Euclidean norms. For example, in Section 4.1, we need to solve subproblems of the form (19), where the distance in the $\mathbf{W}$-space is defined by the $\|\mathbf{W}\|_{\mathbf{M}}$-norm. By an analysis similar to that of Lemma 6 and noting that $\|\mathbf{M}^{-1}\| \leq 1$, we obtain the distance between $\mathbf{w}_i^t$ and the optimal solution $\widetilde{\mathbf{w}}_i^{t+1}$ is at most $L/\gamma$. As a result, the suboptimality of solving (19) when initialized from $\mathbf{W}^t$ is at most $L^2/\gamma$.

# G Directly optimizing the loss with bounded delays

When directly optimizing the loss (while linearizing the regularizer), consider the case where the synchronization step is not perfect. Instead of waiting for neighboring machines to finish their local proximal step and sending in their new weight parameters, each machine can use the stale parameters for neighboring machines. Can we still solve the original ERM problem in this case?

Consider the iteration $t + 1$ on machine $i$ (with delays, $t$ is now considered a local iteration counter). Let the set of neighboring machines be $\mathcal{N}_i$. Due to delay in communication, we have a noisy gradient

$$\widetilde{\nabla}_i R(\mathbf{W}^t) = \frac{1}{m}\left(\eta \mathbf{w}_i^t + \tau \sum_k a_{ik}(\mathbf{w}_i^t - \mathbf{w}_k^{t-d_{ik}(t)})\right), \qquad i = 1, \ldots, m.$$

Here $d_{ik}(t) \in [0, \Gamma]$ is the delay of machine $k$ relative to machine $i$ (at iteration $t + 1$): Machine $i$ is using the weight of machine $k$ from $d_{ik}(t)$ steps ago. In this section, we allow the delay to vary over time, as long as it is upper bounded by $\Gamma$.

Based on this noisy gradient, machine $i$ computes the following proximal gradient step

$$\mathbf{w}_i^{t+1} = \mathrm{prox}_{\frac{\widehat{F}_i}{m}}^{\beta}\left(\mathbf{w}_i^t - \frac{1}{\beta}\widetilde{\nabla}_i R(\mathbf{W}^t)\right) \qquad (20)$$

with some stepsize $\beta > 0$. We need to analyze the convergence of the proximal gradient method with errors in the gradient, as done by Schmidt et al. (2011). The difference from their work is that the error in our gradients comes from delay (stale weight parameters).

Comparing with the case without delay, we have the "error" in the local gradient:

$$\widetilde{\nabla}_i R(\mathbf{W}^t) - \nabla_i R(\mathbf{W}^t) = \frac{\tau}{m} \sum_k a_{ik}(\mathbf{w}_k^t - \mathbf{w}_k^{t-d_{ik}(t)}).$$

From iteration $t - d_{ik}(t)$ to iteration $t$, the $k$-th machine has performed $d_{ik}(t)$ gradient proximal operations. The intuition is that, by the non-expansiveness of the proximal operator, the error in gradient would not cause too much error in the iterates, and then by the smoothness of the objective, this would in turn only results in small error in gradient of the next step. It is important to note that, all machines are influenced by each other and the local errors are propagated to the entire graph.

Based on the non-expansive property of the proximal operator and the additional assumption of the adjacency matrix being doubly-stochastic, it is straightforward to show the following convergence guarantee for the (non-accelerated) proximal gradient algorithm. The algorithm converges at a slower linear rate than without delays.

**Theorem 7.** *Assume that the affinity matrix $\mathbf{A}$ is doubly-stochastic, i.e., $\sum_{k \in \mathcal{N}_i} a_{ik} = 1$ for all $i$, and the delay in the update rule (20) has delay bounded by $\Gamma$. Set the inverse stepsize $\beta = \frac{\eta+\tau}{m}$. Then after $t \geq 1$ iterations of the algorithm, we have*

$$\max_{i=1,\ldots,m} \left\|\mathbf{w}_i^t - \widehat{\mathbf{w}}_i\right\| \leq \left(1 - \frac{\eta}{\eta + \tau}\right)^{\frac{t}{1+\Gamma}} \cdot \max_{i=1,\ldots,m} \left\|\mathbf{w}_i^0 - \widehat{\mathbf{w}}_i\right\|.$$

*Proof.* Since $\widehat{\mathbf{W}}$ is the optimal solution to the ERM problem, we have that

$$\widehat{\mathbf{w}}_i = \text{prox}^{\beta}_{\frac{\widehat{F}_i}{m}} \left( \widehat{\mathbf{w}}_i - \frac{1}{\beta} \nabla_i R(\widehat{\mathbf{W}}) \right), \qquad i = 1, \ldots, m.$$

Then, by the non-expansiveness of the proximal operator, we obtain

$$\begin{aligned}
\|\mathbf{w}_i^{t+1} - \widehat{\mathbf{w}}_i\| &= \left\| \text{prox}^{\beta}_{\frac{\widehat{F}_i}{m}} \left( \mathbf{w}_i^t - \frac{1}{\beta} \widetilde{\nabla}_i R(\mathbf{W}^t) \right) - \text{prox}^{\beta}_{\frac{\widehat{F}_i}{m}} \left( \widehat{\mathbf{w}}_i - \frac{1}{\beta} \nabla_i R(\widehat{\mathbf{W}}) \right) \right\| \\
&\leq \left\| \left( \mathbf{w}_i^t - \frac{1}{\beta} \widetilde{\nabla}_i R(\mathbf{W}^t) \right) - \left( \widehat{\mathbf{w}}_i - \frac{1}{\beta} \nabla_i R(\widehat{\mathbf{W}}) \right) \right\| \\
&= \left\| \left( 1 - \frac{\eta + \tau \sum_{k \in \mathcal{N}_i} a_{ik}}{\beta m} \right) (\mathbf{w}_i^t - \widehat{\mathbf{w}}_i) + \sum_{k \in \mathcal{N}_i} \frac{\tau a_{ik}}{\beta m} (\mathbf{w}_k^{t - d_{ik}(t)} - \widehat{\mathbf{w}}_k) \right\| \\
&\leq \left( 1 - \frac{\eta + \tau \sum_{k \in \mathcal{N}_i} a_{ik}}{\beta m} \right) \|\mathbf{w}_i^t - \widehat{\mathbf{w}}_i\| + \frac{\tau}{\beta m} \sum_{k \in \mathcal{N}_i} a_{ik} \left\| \mathbf{w}_k^{t - d_{ik}(t)} - \widehat{\mathbf{w}}_k \right\| \\
&\leq \left( 1 - \frac{\eta + \tau \sum_{k \in \mathcal{N}_i} a_{ik}}{\beta m} \right) \|\mathbf{w}_i^t - \widehat{\mathbf{w}}_i\| + \frac{\tau}{\beta m} \sum_{k \in \mathcal{N}_i} a_{ik} \max_{t - \Gamma \leq t' \leq t} \left\| \mathbf{w}_k^{t'} - \widehat{\mathbf{w}}_k \right\| \qquad (21)
\end{aligned}$$

where we have used the triangle inequality in the second inequality.

Assume that the affinity matrix $\mathbf{A}$ is doubly-stochastic, so that $\sum_{k \in \mathcal{N}_i} a_{ik} = 1$ for all $i$. Denote $V(t) = \max_{i=1,\ldots,m} \|\mathbf{w}_i^t - \widehat{\mathbf{w}}_i\|$. Then (21) implies that $\|\mathbf{w}_i^{t+1} - \widehat{\mathbf{w}}_i\| \leq \left( 1 - \frac{\eta + \tau}{\beta m} \right) V(t) + \frac{\tau}{\beta m} \max_{t - \Gamma \leq t' \leq t} V(t')$ holds for all $i$, and as a result

$$V(t+1) \leq \left( 1 - \frac{\eta + \tau}{\beta m} \right) V(t) + \frac{\tau}{\beta m} \max_{t - \Gamma \leq t' \leq t} V(t').$$

As long as $\beta \geq \frac{\eta + \tau}{m}$, we have $\left( 1 - \frac{\eta + \tau}{\beta m} \right) \in [0, 1]$. Then according to Feyzmahdavian et al. (2014, Lemma 3), we have

$$V(t) \leq \left( 1 - \frac{\eta}{\beta m} \right)^{\frac{t}{1 + \Gamma}} V(0).$$

Setting $\beta$ to be the smallest possible value $\frac{\eta + \tau}{m}$ yields the desired result. $\square$

## H Comparisons with previous distributed multi-task learning algorithms

We now provide upper bounds of the iteration complexities for the distributed multi-task learning algorithms of Vanhaesebrouck et al. (2017) and Liu et al. (2017) in the ERM setting. We convert their notations into ours to be consistent.

## H.1 Iteration complexity of the algorithm of Liu et al. (2017)

The full algorithm of Liu et al. (2017) performs alternating optimization over the task relationship and the local predictors on each machine. In order to to compare their algorithm with ours on the efficiency of learning predictors, we consider a fixed task correlation matrix $\mathbf{M} = \mathbf{I} + \frac{\tau}{\eta}\mathbf{L}$ in their objective (corresponding to $\Omega$ in eqn (1) of their paper).

With fixed $\mathbf{M}$, their algorithm performs distributed SDCA (Ma et al., 2015) for optimizing over the predictors. In each round of distributed SDCA, one constructs an upper bound of the objective that is separable over the machines (predictors), so that each machine solves a subproblem defined by its local data, and then one around of communication is used to aggregate local updates.

When the instantaneous losses are $\beta_F$-smooth and each local subproblem is solved exactly (i.e., we set $\Theta = 0$ in their analysis), the number of global (communication) rounds needed for obtaining an approximate solution is, according to Liu et al. (2017, Lemma 7 and Theorem 8), of the order (ignoring the logarithmic factor on final optimization error)

$$\max_{\boldsymbol{\alpha}} \frac{\boldsymbol{\alpha}^\top \mathbf{K}\boldsymbol{\alpha}}{\sum_{i=1}^m \boldsymbol{\alpha}_{[i]}^\top \mathbf{K}\boldsymbol{\alpha}_{[i]}} \cdot \max_i \left(\mathbf{M}^{-1}\right)_{ii} \cdot \frac{\beta_F}{\eta}.$$

Here, the first term measures the "task separability" with value in $[1, m]$ (see the definitions of $\mathbf{K}$ and $\boldsymbol{\alpha}_{[i]}$ in their Theorem 1, and the discussion of separability in Section 6.3). On the other hand, we have $\max_i \left(\mathbf{M}^{-1}\right)_{ii} \leq \sigma_{\max}\left(\mathbf{M}^{-1}\right) \leq 1$. As a result, the iteration complexity of distributed SDCA is

$$\widetilde{\mathcal{O}}\left(\frac{\beta_F}{\eta}\right) \times \text{(task separability in [1,m])}.$$

This iteration complexity is similar to that of our ERM algorithm by directly solving the regularizer ($\widetilde{\mathcal{O}}\left(\sqrt{\frac{\beta_F}{\eta}}\right)$), but has worse dependence on the condition number and an unclear multiplicative constant on the tasks separability.

## H.2 Comparison with the collaborative algorithm of Vanhaesebrouck et al. (2017)

We now compare with the collaborative learning algorithm of Vanhaesebrouck et al. (2017) in the synchronous and decentralized setting. In their algorithm, each machine augments its local optimization parameters to include a copy of predictor from each neighboring machine. Let $\Theta_i$ be the set of $|\mathcal{N}_i| + 1$ variables $\mathbf{w}_k$ for $k \in \mathcal{N}_i \cup \{i\}$, and $\Theta_i^k$ is the copy of $\mathbf{w}_k$ on machine $i$. We can reformulate the global objective (2) as

$$\arg\min_{\{\Theta_i\}_{i=1}^m} \sum_{i=1}^m H_i(\Theta_i) \quad \text{where } H_i(\Theta_i) = \frac{1}{m}\widehat{F}_i(\Theta_i^i) + \frac{\eta}{2m}\left\|\Theta_i^i\right\|^2 + \frac{\tau}{4m}\sum_{k \in \mathcal{N}_i} a_{ik}\left\|\Theta_i^i - \Theta_i^k\right\|^2$$

$$\text{subject to} \quad \Theta_i^i = \Theta_k^i, \quad \text{for all } (i, k) \text{ s.t. } k \in \mathcal{N}_i. \tag{22}$$

Vanhaesebrouck et al. (2017) then introduce variables associated with each edge (4 set of variables per edge) and apply ADMM to the resulting problem. An advantage of ADMM is that it allows decoupling of the local problems when updating primal variables, where the local problem involves the nonlinearized loss function.

Although Vanhaesebrouck et al. (2017) suggest that the convergence results of synchronous decentralized ADMM (Wei and Ozdaglar, 2013; Shi et al., 2014) apply to this formulation (see their Appendix D), we note however that (22) is not in the standard form covered by these results. In particular, the classical decentralized concensus problem has the form

$$\min_{\mathbf{x}_1,\ldots,\mathbf{x}_m} \sum_{i=1}^m f_i(\mathbf{x}_i) \quad \text{s.t.} \quad \mathbf{x}_i = \mathbf{x}_j \quad \text{for all } (i,j) \text{ where } j \in \mathcal{N}_i.$$

Here, neighboring machines share the same set of optimization parameters and they would like to reach complete consensus, whereas in (22) neighboring machines can have different set of variables and they only try to achieve consensus on the shared parameters. As a result, it is nontrivial to derive the iteration complexity of the collaborative learning algorithm of Vanhaesebrouck et al. (2017) based on the same quantities used in the analysis of our algorithms.

## I  Experiments

In this section we examine the empirical performance of the proposed algorithms. We consider the problem of linear regression on synthetic data. For the $i$-th task, we generate data from

$$y = \langle \mathbf{w}_i^*, \mathbf{x} \rangle + \epsilon,$$

where $\epsilon$ is noise drawn from the Normal distribution $\mathcal{N}(0,3)$, $\mathbf{x} \in \mathbb{R}^d$ is drawn from a multivariate Normal distribution with mean zero and covariance matrix $\mathbf{\Sigma}$ where $\mathbf{\Sigma}_{ij} = 2^{-|i-j|/3}$, and $\mathbf{w}_i^* \in \mathbb{R}^d$ is a coefficient vector for the $i$-th task generated from the following clustered multi-task structure. Each $\mathbf{w}_i^*$ is drawn from a mixture of $C$ clusters; there is a reference model $\mathbf{r}_j$ for each cluster $j = 1, \ldots, C$, and the task specific model $\mathbf{w}_i^*$ is a small perturbation of the corresponding cluster reference model:

$$\mathbf{w}_i^* = \mathbf{r}_j + \xi_i, \quad \text{if} \quad \mathbf{w}_i^* \text{ is drawn from cluster } j.$$

The cluster reference model $\mathbf{r}_j$ is generated by sampling each entry i.i.d. from $Unif[-0.5, 0.5]$, while the perturbation vector $\xi_i$ is generated by sampling each entry i.i.d. from $Unif[-0.05, 0.05]$. This construction gives us task specific models which are similar to each other when they belong to the same cluster. The corresponding similarity graph is a 10-nearest neighbor graph (so the graph is connected) with binary weights built on $\{\mathbf{w}_i\}_{i=1,\ldots,m}$, i.e., each task is connected to 10 other tasks whose models are most similar.

We tested a few graph-based multi-task learning methods.

- `Local`: solves a local ERM problem (with only $\ell_2$ regularization) with $n$ samples for each task.

- `Centralized`: solves the graph-regularized ERM problem (2) with $n$ samples for each task.

- `ADMM`: the synchronized version of the ADMM algorithm of Vanhaesebrouck et al. (2017).

- `SDCA`: the distributed SDCA algorithm of Liu et al. (2017) for fixed graph.

- Our algorithms: denoted as B/S (batch/stochastic) + SR/OL (solve regularizer/optimize loss).

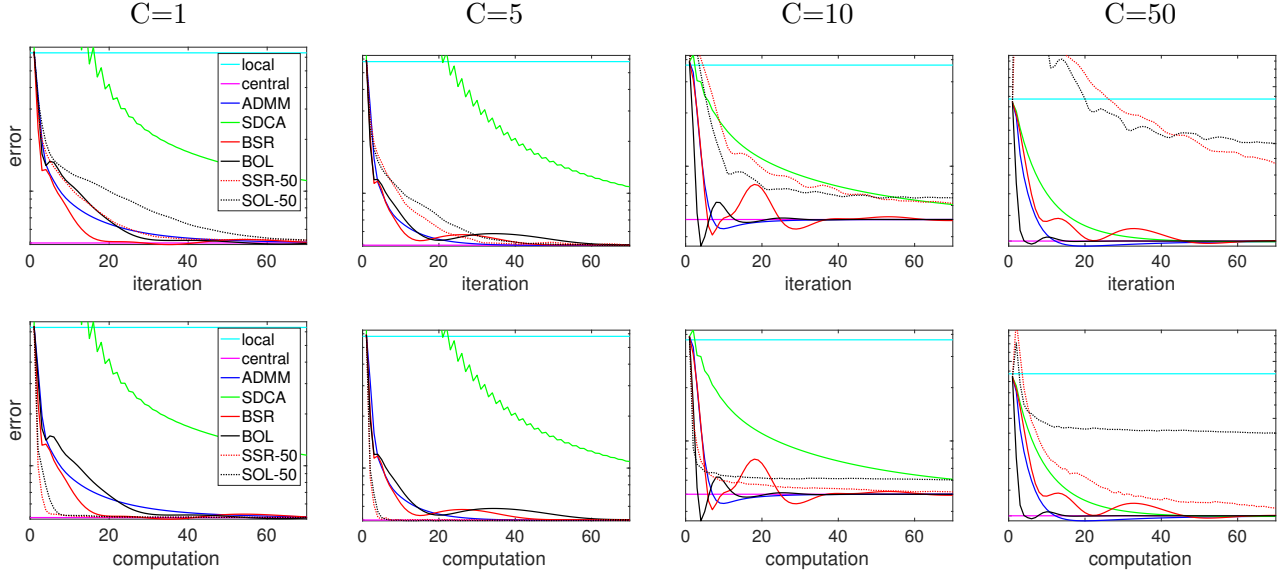

Figure 2: Performance of different methods for regularized empirical risk minimization.

In the experiments below, we have problem dimension $d = 100$, number of tasks $m = 100$, training set size $n = 500$, and vary number of task clusters $C$ over $\{1, 5, 10, 50\}$ (smaller $C$ implies overall stronger task similarity). We also generate a dev set of 10000 samples per task for tuning hyper-parameters, and test set of 10000 samples per task for approximately evaluating the population loss.

**Empirical risk minimization** We fist compare the iterative methods on the regularized ERM problem (2), to which the analysis for `ADMM` and `SDCA` applies. We tune the $\ell_2$ regularization parameter for `Local` and $(\eta, \tau)$ for `Centralized`, and then fix the optimal $(\eta, \tau)$ for other methods. We also tune the quadratic penalty parameter for `ADMM`, the task separability and stepsize parameters for `SDCA`, and stepsize parameter for BSR/BOL (although the default value based on the smoothness parameter already works well for them). For SSR/SOL, we draw random samples from the fixed training set (with size $n$), and simply fix the minibatch size to be $n/10$.

Figure 2 shows for each method the estimated $F(\mathbf{W})$ over iterations (or rounds of communication) in the top row, and over the amount of computation (measured by the number of passes over the training set) in the bottom row. Observe that all iterative algorithms converge to the same ERM solution, our algorithms tend to consistently outperform `ADMM` and `SDCA`.

**Stochastic optimization** We next demonstrate the efficiency of true stochastic algorithms (using fresh samples for each update) at $C = 10$. We allow the algorithms to process a total of 10000 fresh samples on each machine, and vary the minibatch size $b$ over $\{40, 80, 100, 200, 500\}$. The parameters $(\eta, \tau)$ are fixed to those used in the ERM experiments.

Figure 3 shows for each method the estimated $F(\mathbf{W})$ over iterations (or rounds of communication) in the left plot, and over the amount of fresh samples processed (or total computation cost) in the right plot. As a reference, the error of `Local` and `Centralized` (using $n = 500$ samples per machine) are also given in the plots. We observe that with fresh samples, stochastic algorithms are competitive to ERM algorithms in terms of sample complexity, while being computationally more efficient.

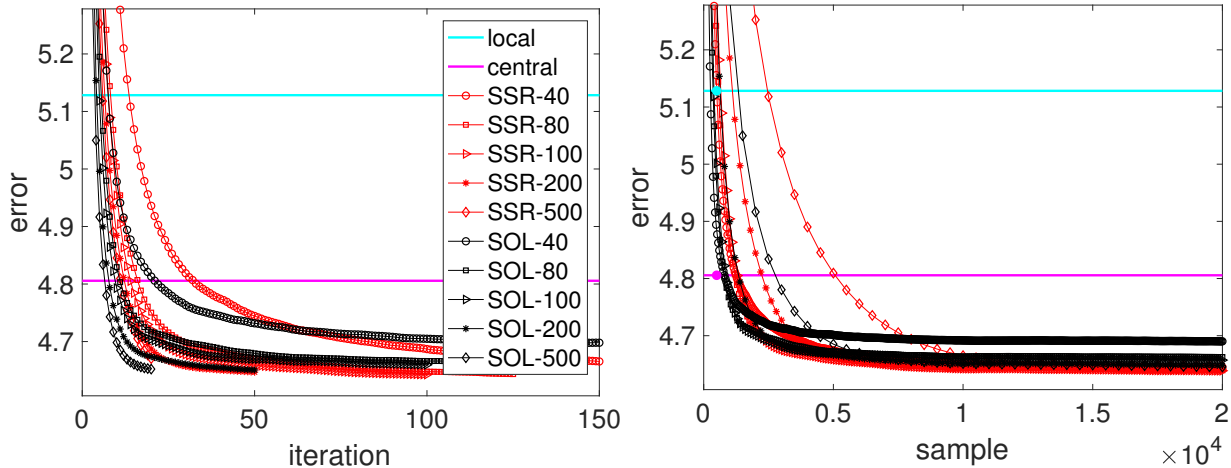

Figure 3: Performance of stochastic algorithms with various minibatch sizes. Here $C = 10$.

# References

Y. Amit, M. Fink, N. Srebro, and S. Ullman. Uncovering shared structures in multiclass classification. In *Proceedings of the 24th international conference on Machine learning*, pages 17–24. ACM, 2007.

R. K. Ando and T. Zhang. A framework for learning predictive structures from multiple tasks and unlabeled data. *J. Mach. Learn. Res.*, 6:1817–1853, 2005.

A. Argyriou, T. Evgeniou, and M. Pontil. Convex multi-task feature learning. *Mach. Learn.*, 73 (3):243–272, 2008.

M.-F. Balcan, A. Blum, S. Fine, and Y. Mansour. Distributed learning, communication complexity and privacy. In S. Mannor, N. Srebro, and R. Williamson, editors, *JMLR W&CP 23: COLT 2012*, volume 23, pages 26.1–26.22, 2012.

I. M. Baytas, M. Yan, A. K. Jain, and J. Zhou. Asynchronous multi-task learning. In *IEEE International Conference on Data Mining (ICDM)*, 2016.

S. P. Boyd, N. Parikh, E. Chu, B. Peleato, and J. Eckstein. Distributed optimization and statistical learning via the alternating direction method of multipliers. *Found. Trends Mach. Learn.*, 3(1): 1–122, 2011.

O. Dekel, R. Gilad-Bachrach, O. Shamir, and L. Xiao. Optimal distributed online prediction using mini-batches. *Journal of Machine Learning Research*, 13:165–202, 2012.

T. Evgeniou and M. Pontil. Regularized multi-task learning. In *Proceedings of the tenth ACM SIGKDD international conference on Knowledge discovery and data mining*, pages 109–117. ACM, 2004.

T. Evgeniou, C. A. Micchelli, and M. Pontil. Learning multiple tasks with kernel methods. In *J. Mach. Learn. Res.*, pages 615–637, 2005.

H. R. Feyzmahdavian, A. Aytekin, and M. Johansson. A delayed proximal gradient method with linear convergence rate. In *2014 IEEE International Workshop on Machine Learning for Signal Processing (MLSP)*, 2014.

R. Johnson and T. Zhang. Accelerating stochastic gradient descent using predictive variance reduction. In *Advances in Neural Information Processing Systems (NIPS)*, pages 315–323, 2013.

J. Konecny, B. McMahan, and D. Ramage. Federated optimization: Distributed optimization beyond the datacenter. arXiv:1511.03575 [cs.LG], 2015.

G. Lan. An optimal method for stochastic composite optimization. *Math. Prog.*, 133(1–2):365–397, 2012.

S. Liu, S. J. Pan, and Q. Ho. Distributed multi-task relationship learning. arXiv:1612.04022 [cs.LG], 2017.

K. Lounici, M. Pontil, A. B. Tsybakov, and S. A. van de Geer. Oracle inequalities and optimal inference under group sparsity. *Ann. Stat.*, 39:2164–204, 2011.

C. Ma, V. Smith, M. Jaggi, M. I. Jordan, P. Richtarik, and M. Takac. Adding vs. averaging in distributed primal-dual optimization. In *Proc. of the 32st (ICML 2015)*, 2015.

A. Maurer. The Rademacher complexity of linear transformation classes. In G. Lugosi and H.-U. Simon, editors, *Annual Conference on Learning Theory*, pages 65–78, 2006.

A. Nedić and A. Ozdaglar. Distributed subgradient methods for multi-agent optimization. *IEEE Trans. Automat. Contr.*, 54(1):48–61, 2009.

Y. Nesterov. *Introductory Lectures on Convex Optimization. A Basic Course.* Number 87. Springer-Verlag, 2004.

G. Obozinski, M. J. Wainwright, and M. I. Jordan. Support union recovery in high-dimensional multivariate regression. *Ann. Stat.*, 39(1):1–47, 2011.

S. S. Ram, A. Nedić, and V. V. Veeravalli. Distributed stochastic subgradient projection algorithms for convex optimization. *Journal of optimization theory and applications*, 147(3):516–545, 2010.

K. Scaman, F. Bach, S. Bubeck, Y. T. Lee, and L. Massoulie. Optimal algorithms for smooth and strongly convex distributed optimization in networks. arXiv:1702.08704 [math.OC], 2017.

M. Schmidt, N. L. Roux, and F. Bach. Convergence rates of inexact proximal-gradient methods for convex optimization. pages 1458–1466, 2011.

S. Shalev-Shwartz, O. Shamir, N. Srebro, and K. Sridharan. Stochastic convex optimization. In S. Dasgupta and A. Klivans, editors, *Proc. of the 22th Annual Conference on Learning Theory (COLT'09)*, Montreal, Quebec, 2009.

O. Shamir and N. Srebro. Distributed stochastic optimization and learning. In *52nd Annual Allerton Conference on Communication, Control, and Computing (Allerton), 2014*, pages 850–857. IEEE, 2014.

W. Shi, Q. Ling, K. Yuan, G. Wu, and W. Yin. On the linear convergence of the ADMM in decentralized consensus optimization. *IEEE Trans. Signal Processing*, 62(7):1750–1761, 2014.

B. A. Turlach, W. N. Venables, and S. J. Wright. Simultaneous variable selection. *Technometrics*, 47(3):349–363, 2005.

P. Vanhaesebrouck, A. Bellet, and M. Tommasi. Decentralized collaborative learning of personalized models over networks. In *Int. Workshop on Artificial Intelligence and Statistics*, pages 509–517, 2017.

J. Wang, M. Kolar, and N. Srebro. Distributed multitask learning. *ArXiv e-prints, arXiv:1510.00633*, 2015, `arXiv:1510.00633`.

J. Wang, M. Kolar, and N. Srebro. Distributed multi-task learning with shared representation. 2016, `arXiv:1603.02185`.

J. Wang, W. Wang, and N. Srebro. Memory and communication efficient distributed stochastic optimization with minibatch prox. In S. Kale and O. Shamir, editors, *Annual Conference on Learning Theory*, Amsterdam, Netherlands, 2017.

E. Wei and A. Ozdaglar. On the $o(1/k)$ convergence of asynchronous distributed alternating direction method of multipliers. arXiv:1307.8254 [math.OC], 2013.

K. Yuan, Q. Ling, and W. Yin. On the convergence of decentralized gradient descent. *SIAM Journal on Optimization*, 26(3):1835–1854, 2016.

M. Yuan, A. Ekici, Z. Lu, and R. Monteiro. Dimension reduction and coefficient estimation in multivariate linear regression. *J. R. Stat. Soc. B*, 69(3):329–346, 2007.